\documentclass[sigconf]{acmart}
\usepackage{graphicx} 
\usepackage{multirow}
\usepackage{tabu}
\usepackage{bm}

\usepackage{algorithm}
\usepackage{algpseudocode}

\usepackage{booktabs}
\usepackage{threeparttable}
\usepackage{xcolor}
\usepackage{subfigure}

\usepackage{balance}

\usepackage{amsmath}
\usepackage{amssymb}

\AtBeginDocument{%
  \providecommand\BibTeX{{%
    \normalfont B\kern-0.5em{\scshape i\kern-0.25em b}\kern-0.8em\TeX}}}

\copyrightyear{2022} 
\acmYear{2022} 
\setcopyright{acmcopyright}
\acmConference[MM '22] {Proceedings of the 30th ACM International Conference on Multimedia }{October 10--14, 2022}{Lisboa, Portugal.}
\acmBooktitle{Proceedings of the 30th ACM International Conference on Multimedia (MM '22), Oct. 10--14, 2022, Lisboa, Portugal}
\acmPrice{15.00}
\acmISBN{978-1-4503-9203-7/22/10}
\acmDOI{10.1145/3503161.3548158}

\begin{document}

\title{Few-shot Image Generation Using Discrete Content Representation}

\acmSubmissionID{1685}





\settopmatter{authorsperrow=4}
\author{Yan Hong}
\affiliation{%
  \institution{MoE Key Lab of Artificial Intelligence, Shanghai Jiao Tong University}
  \country{China}
}
\email{yanhong.sjtu@gmail.com}

\author{Li Niu}
\authornote{Corresponding author.}
\affiliation{%
  \institution{MoE Key Lab of Artificial Intelligence, Shanghai Jiao Tong University}
    \country{China}
}
\email{ustcnewly@sjtu.edu.cn}

\author{Jianfu Zhang}
\affiliation{%
  \institution{MoE Key Lab of Artificial Intelligence, Shanghai Jiao Tong University}
  \country{China}
}
\email{c.sis@sjtu.edu.cn}



\author{Liqing Zhang}
\affiliation{\institution{MoE Key Lab of Artificial Intelligence, Shanghai Jiao Tong University}
\country{China}
}
\email{zhang-lq@cs.sjtu.edu.cn}

%


\renewcommand{\shortauthors}{Yan Hong, Li Niu, Jianfu Zhang, \& Liqing Zhang}

\begin{abstract}
Few-shot image generation and few-shot image translation are two related tasks, both of which aim to generate new images for an unseen category with only a few images. In this work, we make the first attempt to adapt few-shot image translation method to few-shot image generation task. Few-shot image translation disentangles an image into style vector and content map. An unseen style vector can be combined with different seen content maps to produce different images. However, it needs to store seen images to provide content maps and the unseen style vector may be incompatible with seen content maps. To adapt it to few-shot image generation task, we learn a compact dictionary of local content vectors via quantizing continuous content maps into discrete content maps instead of storing seen images. Furthermore, we model the autoregressive distribution of discrete content map conditioned on style vector, which can alleviate the incompatibility between content map and style vector. Qualitative and quantitative results on three real datasets demonstrate that our model can produce images of higher diversity and fidelity for unseen categories than previous methods.
\end{abstract}

\ccsdesc[500]{Computing methodologies~Image representations}
\ccsdesc[300]{Computing methodologies~Neural networks}
\ccsdesc[100]{Computing methodologies~Learning latent representations}

\keywords{Few-shot Learning; Image Generation; Quantization}

\maketitle
\section{Introduction}
In recent years, deep generative models \cite{brock2018large,stylegan1,stylegan2,razavi2019generating} can generate high-quality images but require amounts of training data. However, some long-tail categories or newly emerging categories only have limited data \cite{hong2020f2gan}. There is a risk of overfitting to train or finetune generative models on a small amount of data \cite{antoniou2017data,bartunov2018few}. Therefore, it is imperative to consider training on seen categories with sufficient training images and adapting to an unseen category with only a few images. This task is referred to as few-shot image generation \cite{clouatre2019figr,hong2020matchinggan, hong2020f2gan, hong2020deltagan}, which transfers knowledge from seen categories to generate diverse and realistic images for unseen categories. 
\begin{figure}
\begin{center}
\includegraphics[scale=0.18]{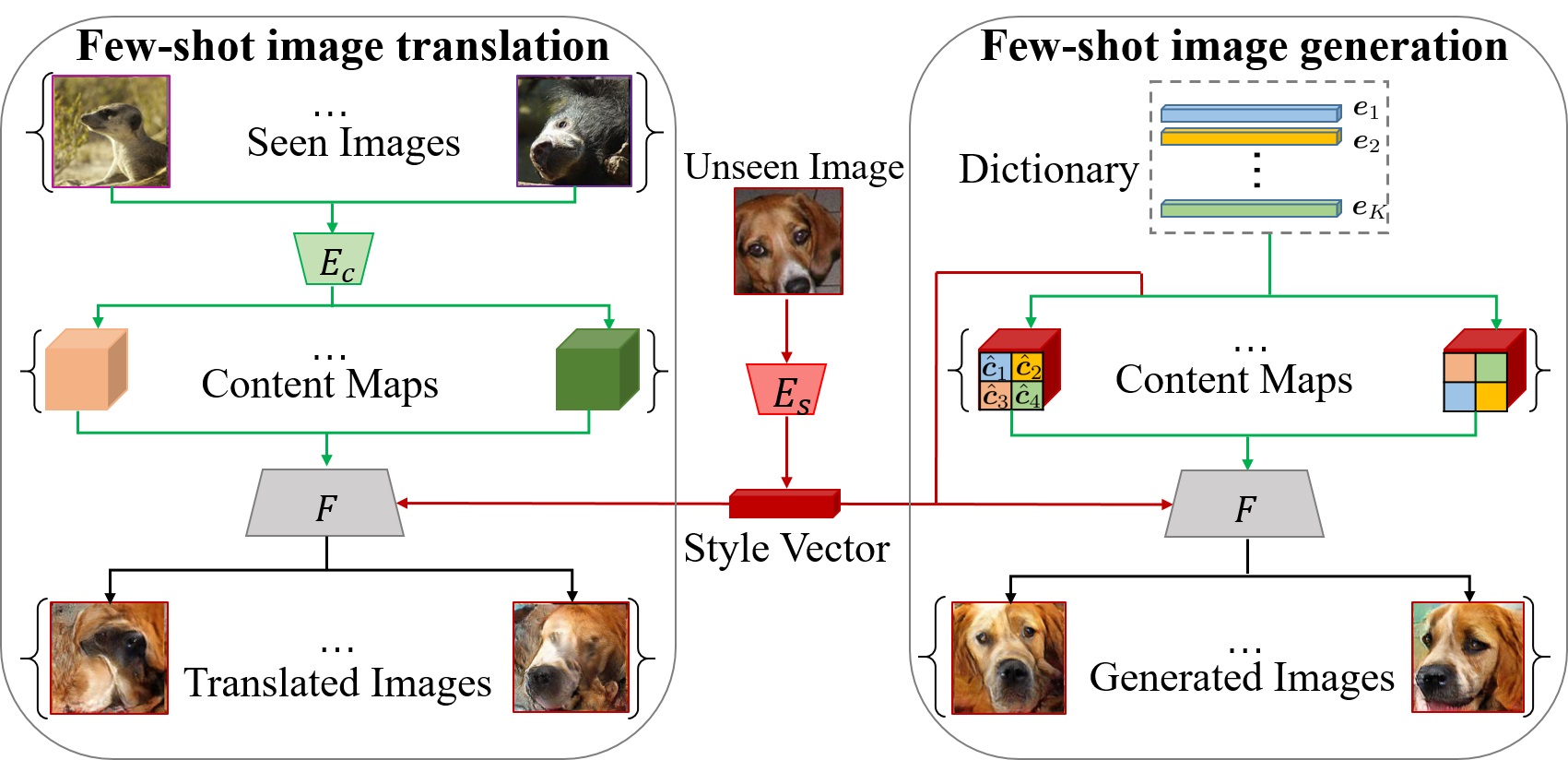}
\end{center}
\caption{Comparison between few-shot image translation and few-shot image generation at test time. The former requires seen images to provide content maps, while the latter samples local content vectors autoregressively conditioned on the style vector to form content maps. }
\label{fig:motivation} 
\end{figure}

In the remainder of this paper, the images from seen (\emph{resp.}, unseen) categories are dubbed as seen (\emph{resp.}, unseen) images for brevity. 
The existing few-shot image generation methods can be roughly classified into optimization-based methods, fusion-based methods, and transformation-based methods. The optimization-based approaches FIGR \cite{clouatre2019figr} and DAWSON \cite{liang2020dawson} finetuned the unconditional generative adversarial network (GAN) for each unseen category with meta-learning algorithms, but the generated unseen images are unrealistic. 
In contrast, fusion-based and transformation-based methods are built upon conditional GAN, which takes one or several conditional images from the same category and generates new images for this category. They reduce the burden of finetuning and support instant adaptation to unseen categories \cite{bartunov2018few}.
For fusion-based methods, GMN \cite{bartunov2018few}, MatchingGAN \cite{hong2020matchinggan}, and F2GAN \cite{hong2020f2gan} fused multiple conditional images with matching procedure to produce new images. However, the generated images are similar to conditional images. For transformation-based methods, DAGAN \cite{antoniou2017data} and DeltaGAN \cite{hong2020deltagan} combined random vectors with conditional image to produce new images. However, DAGAN \cite{antoniou2017data} failed to produce diverse images while DeltaGAN \cite{hong2020deltagan} may generate undesired unseen images due to the mismatch between random vector and conditional image. 

As a closely related task to few-shot image generation, few-shot image translation \cite{liu2019few,saito2020coco} focuses on translating a seen image to an unseen image. 
Previous few-shot image translation works like FUNIT \cite{liu2019few}  disentangle the latent representation into \emph{category-invariant content map} (\emph{e.g.}, pose) and \emph{category-specific style vector} (\emph{e.g.}, appearance). In the testing stage, given an unseen image, its style vector can be combined with the content maps of seen images to produce more unseen images for this unseen category, which is equivalent to translating seen images to this unseen category. 

In this work, we make the first attempt to adapt representative few-shot image translation method FUNIT to few-shot image generation by solving two critical issues. The first issue is that FUNIT relies on the seen images to provide content maps. It is resource-consuming to store a large number of seen images or content maps of seen images. The number of content maps is also limited by the number of seen images. Inspired by VQVAE \cite{van2017neural} which explores discrete representations for compression, we divide the content maps of seen images into local content vectors, in which each local content vector represents the content information of each local part in the image, and compress continuous local content vectors into discrete local content vectors via vector quantization (see Figure \ref{fig:motivation}). In this way, we only need to save a dictionary of local content vectors instead of saving seen images or content maps of seen images. Besides, quantizing content maps can help prune noisy and redundant information \cite{van2017neural}, 
which may benefit the disentanglement of content map and style vector (see Section \ref{sec:Qualitative}). Then, we model the autoregressive distribution of local content vectors, so that we can sample local content vectors autoregressively to form a significant amount of content maps. 
The second issue of FUNIT is that the randomly selected content maps for an unseen image may be incompatible with the style vector,
 resulting in poor translation results as shown in Figure \ref{fig:motivation}. To solve the incompatibility problem, we aim to find compatible content maps for a given unseen image. 
Specifically, we extend the abovementioned autoregressive distribution to conditional autoregressive distribution, which is autoregressive distribution of local content vectors conditioned on a style code. 
In this way, given an unseen image, we can sample local content vectors autoregressively conditioned on its style vector, and the obtained content maps are expected to be compatible with the style vector. Since we extend FUNIT with \textbf{dis}crete \textbf{co}ntent representation, we name our method Disco-FUNIT. 
Our contributions can be summarized as follows: 1) We make the first attempt to adapt few-shot image translation to few-shot image generation, which bridges the gap between two research fields; 2) To avoid the reliance on seen images in the testing stage, we explore discrete local content vectors under the disentanglement framework; 3) To solve the incompatibility issue between style vector and content map, we propose to model conditional autoregressive distribution; 4) Our method can produce diverse and realistic images based on a single image, exceeding existing few-shot image generation methods.


\section{Related Work}
\label{sec:related}


\noindent\textbf{Few-shot Image Generation:}
As the earliest few-shot image generation methods, Bayesian-based methods~\cite{lake2011one,rezende2016one-shot} were proposed to generate new images for simple concepts in few-shot setting. 
Recently, deep learning based few-shot image generation methods can be roughly classified into optimization-based methods~\cite{clouatre2019figr,liang2020dawson} fusion-based methods~\cite{bartunov2018few,hong2020matchinggan,hong2020f2gan}, and transformation-based methods~\cite{antoniou2017data,hong2020deltagan}. Our method can be categorized as the transformation-based method, because we can transform an unseen image to more unseen images from the same category.

Note that some more recent works~\cite{ojha2021few,li2020few,robb2020few,WangGBHK020}
focused on adapting the generative model pretrained on a large dataset to a small dataset with a few examples, which are also called few-shot image generation. However, their setting is quite different from ours. Firstly, these methods adapt from one source domain to another target domain, while our method adapts from multiple seen categories to unseen categories. Secondly, they need to finetune the model for each unseen domain, which is very tedious. Instead, we can make instant adaptation to unseen categories without finetuning.

\noindent\textbf{Few-shot Image Translation:}
In~\cite{benaim2018one}, one image of unseen category is regarded as an exemplar to guide image translation.
FUNIT~\cite{liu2019few} utilized a disentanglement framework to combine the content maps of seen images with the style vectors of unseen images to produce translated images belonging to unseen categories. Built upon FUNIT, COCO-FUNIT~\cite{saito2020coco} proposed a content-conditioned style encoder architecture to adjust the style code. SEMIT~\cite{wang2020semi} proposed a semi-supervised method for few-shot image translation by
reducing the amount of required labeled data during training. In this work, we build upon FUNIT and adapt it to few-shot image generation. 

\noindent\textbf{Vector Quantization:}
The feature vectors learned from convolutional network always contain a lot of redundant information~\cite{wallace1992jpeg,hu2020coarse,fan2020scale}. To compress these learned features, vector quantization methods~\cite{van2017neural} have been proposed to construct a dictionary of discrete vectors to approximate the actual continuous vectors. In other words, one original continuous vector can be represented by its nearest center in the dictionary. Vector quantization can significantly reduce the size of the storage footprint. In this work, we utilize the vector quantization technique to create a dictionary of local content vectors, which serves as a foundation of modeling the compatibility between content map and style vector.

\noindent\textbf{Autoregressive Models:}
The autoregressive models are a type of generative models that have been widely used for image generation, in which the autoregressive distribution can be modeled by different sequence-to-sequence models like PixelCNN, LSTM, and Transformer. 
PixelCNN~\cite{oord2016conditional} modeled the distribution of a natural image using the elementary chain rule over pixels. LSTM~\cite{sutskever2014sequence} exploited the hierarchical relationships between complete images and image portions. Transformer modeled the long-range interactions between local image representations~\cite{parmar2018image,weissenborn2019scaling}. In this work, we leverage Transformer to model the autoregressive distribution.




\begin{figure*} [t]
\begin{center}
\includegraphics[scale=0.3]{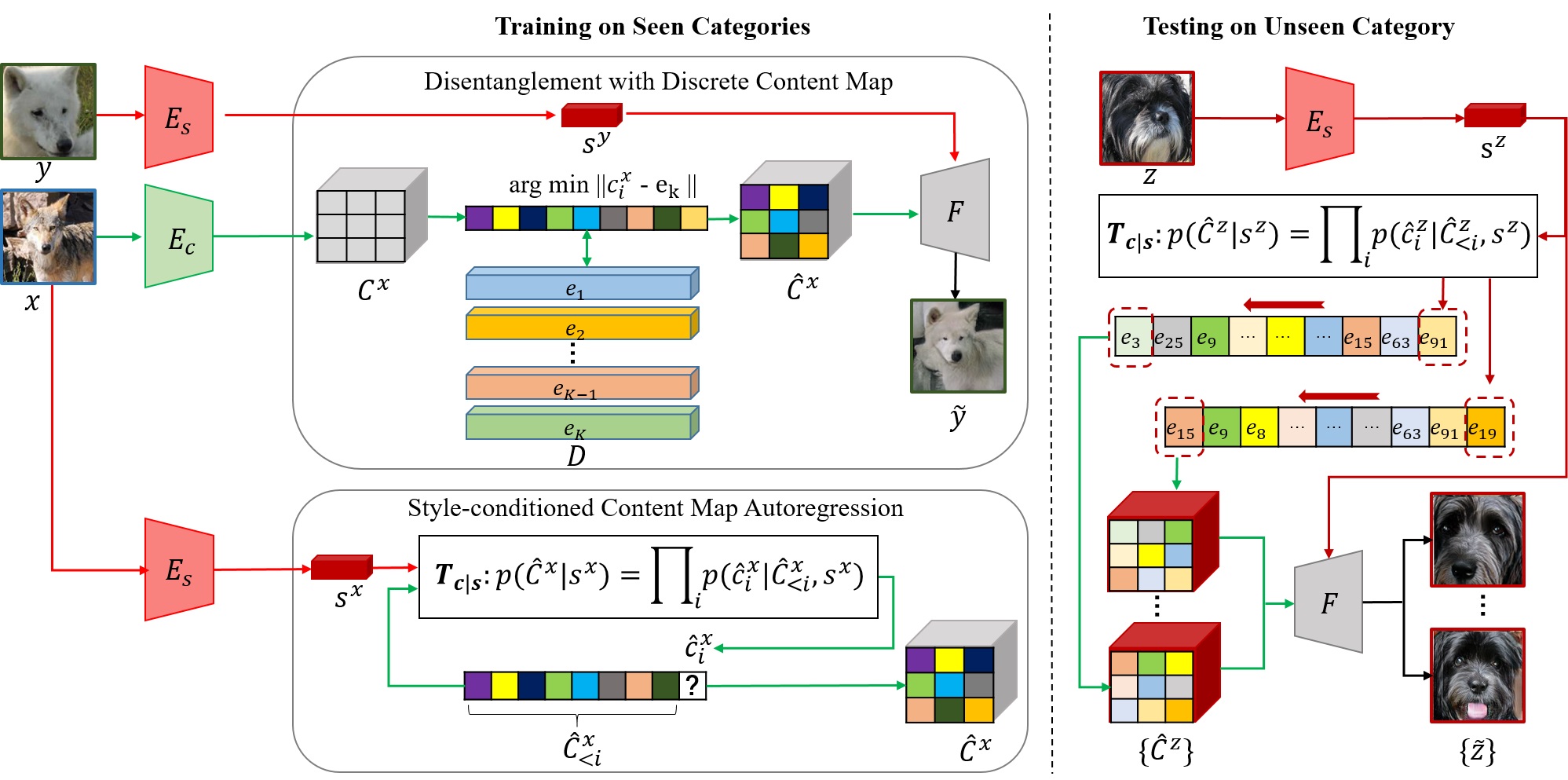}
\end{center}
\caption{Our method consists of two training stages: disentanglement with discrete content map and style-conditioned content map autoregression. In the first stage, we jointly learn an FUNIT model and a dictionary $\mathcal{D}$, during which continuous content map $\bm{C}^x$ is quantized into discrete content map $\hat{\bm{C}}^x$. In the second stage, we learn a content map generator $T_{c|s}$ to model the autoregressive distribution of $\hat{\bm{C}}^x$ conditioned on the style vector $\bm{s}^x$. At test time, given a unseen image $\bm{z}$, $T_{c|s}$ samples diverse content maps $\{\hat{\bm{C}}^z\}$ autoregressively conditioned on the style vector $\bm{s}^z$ to produce diverse images $\{\tilde{\bm{z}}\}$.
} 
\label{fig:framework} 
\end{figure*}

\section{Background on FUNIT}\label{sec:funit}
Since our goal is adapting few-shot image translation method FUNIT~\cite{liu2019few} to few-shot image generation, we first have a brief introduction to FUNIT, which is basically a disentanglement network to disentangle content map and style  vector.
All categories are split into non-overlapping seen categories and unseen categories. FUNIT is trained on seen categories. Then, the learned model can leverage an image of an unseen category to translate any seen image to this unseen category.

FUNIT consists of a generator $G$ and a multi-class discriminator $D$, where $G$ is constructed with a content encoder $E_c$, a style encoder $E_s$, and a decoder $F$. 
During training, we randomly sample an image $\bm{x}$ from a seen category $l^x$ and an image $\bm{y}$ from another seen category $l^y$. We use $E_c$ to extract content map $\bm{C}^x \in \mathbb{R}^{w\times h\times {d_c}}$ from $\bm{x}$: $\bm{C}^x =E_c(\bm{x})$, where $w\times h$ (\emph{resp.}, $d_c$) is the spatial size (\emph{resp.}, channel dimension) of content map. 
We use $E_s$ to extract style vector $\bm{s}^{y} \in \mathbb{R}^{{d_s}}$ from  $\bm{y}$: $\bm{s}^y =E_s(\bm{y})$, where $d_s$ is the dimension of style vector. 
Then, content map $\bm{C}^x$ and style vector $\bm{s}^y$ are fed into the decoder $F$ to produce translated image $\tilde{\bm{y}}$:
\begin{equation}\label{eqn:quantised_translation}
\begin{aligned}
\tilde{\bm{y}} = F(\bm{C}^x, \bm{s}^y).
\end{aligned}
\end{equation}
$\tilde{\bm{y}}$ integrates the content information of $\bm{x}$ and the style information of $\bm{y}$. Because style is category-specific information, $\tilde{\bm{y}}$ belongs to the same category as $\bm{y}$.
The multi-class discriminator $D$ is designed to ensure the realism of translated image $\tilde{\bm{y}}$ given the category label $l^y$ by optimizing adversarial loss:
\begin{eqnarray}
\!\!\!\!\!\!\!\!&&\mathcal{L}_D = \mathbb{E}_{\tilde{\bm{y}}} [\max (0,1+\mathrm{D}^{y}(\tilde{\bm{y}})]  +  \mathbb{E}_{\bm{y}}  [\max (0,1-\mathrm{D}^{y}({\bm{y}}))], \nonumber\\
\!\!\!\!\!\!\!\!&&\mathcal{L}_{GD} = - \mathbb{E}_{\tilde{\bm{y}}} [\mathrm{D}^{y}(\tilde{\bm{y}})],
\end{eqnarray}
where $D^x$ (\emph{resp.}, $D^y$) denotes the discriminator corresponding to the category $l^x$ (\emph{resp.}, $l^y$). 

When applying $E_c$ and $E_s$ to the same image $\bm{x}$, $F$ is encouraged to reconstruct the input image $\bm{x}$ with a reconstruction loss $\mathcal{L}_{\mathrm{R}} = \mathbb{E}_{\bm{x}} ||\bm{x} - F(\bm{C}^x, \bm{s}^x) ||_1$, where $\bm{s}^x = E_s(\bm{x})$.

To stabilize the adversarial training, feature matching loss $\mathcal{L}_{\mathrm{FM}} = \mathbb{E}_{\bm{y}, \tilde{\bm{y}}} ||\bar{D}({\bm{y}}) - \bar{D}({\bm{\tilde{y}}}) ||_1 $ is designed to minimize the distance between  $\bar{D}({\bm{y}})$ and $\bar{D}(\tilde{\bm{y}})$, where $\bar{D}$ is feature extractor constructed by removing the last layer from $D$. The overall objective of FUNIT is summarized as follows,
\begin{equation}\label{eqn:funit}
\begin{aligned}
\mathcal{L}_{fun} = 
\mathcal{L}_D + \mathcal{L}_{GD} +\lambda_{\mathrm{R}} \mathcal{L}_{\mathrm{R}}+\lambda_{\mathrm{F}} \mathcal{L}_{\mathrm{FM}},
\end{aligned}
\end{equation}
in which $\lambda_{\mathrm{R}}$ and $\lambda_{\mathrm{F}}$ are hyper-parameters.
After training, an input image is disentangled into content map and style vector. In the inference stage, given an image $\bm{z}$ from an unseen category $l^z$, the style vector of $\bm{z}$ can be combined with the content map of any seen image, yielding a new image with category label $l^z$.

\section{Our Method}  \label{sec:method}

In this section, we introduce the process of adapting FUNIT to few-shot image generation. 
As shown in Figure~\ref{fig:framework}, our few-shot image generation method consists of two training stages. The first stage is disentanglement with discrete content map and the second stage is style-conditioned content map autoregression.We refer to the image providing content (\emph{resp.}, style) as content (\emph{resp.}, style) image. In the first stage, given images $\bm{x}$ and $\bm{y}$ from two seen categories, we train an FUNIT model as in Section~\ref{sec:funit},  but the content map $\bm{C}^x$ is quantized into discrete content map $\hat{\bm{C}}^x$ via vector quantization method~\cite{van2017neural}. In the second stage, conditioned on the style vector $\bm{s}^x$ of $\bm{x}$, we model the compatibility between style vector $\bm{s}^x$ and discrete content map $\hat{\bm{C}}^x$ using a content map generator $T_{c|s}$. 
At test time, given an image $\bm{z}$ from unseen category $l^z$, our content map generator $T_{c|s}$ can sample local content vectors autoregressively to form diverse content maps $\hat{\bm{C}}^z$ which are compatible with the style vector $\bm{s}^z$. Then, different content maps $\{\hat{\bm{C}}^z\}$ can be combined with $\bm{s}^z$  to produce diverse and realistic images $\{\tilde{\bm{z}}\}$ from unseen category $l^z$.



\subsection{Disentanglement with Discrete  Content  Map}  \label{sec:stage1}
Based on FUNIT, given an image $\bm{x}$, content encoder $E_c$ extracts its content map $\bm{C}^x \in \mathbb{R}^{w\times h\times {d_c}}$. To avoid the heavy storage cost of content maps and remove the redundant information from content maps, we divide a content map to local content vectors and quantize them into discrete local content vectors. 

Specifically, we spatially split the content map into local content vectors, with each local content vector representing the content information of a local part (\emph{e.g.}, eye, nose) in the original image. Similar to~\cite{kolesnikov2017pixelcnn}, content map $\bm{C}^x$ is ordered from left to right and from top to bottom to form a sequence $\{\bm{c}_i^x|_{i=1}^N\}$, in which $N=w\times h$ and each $\bm{c}_i^x$ is a $d_c$-dim local content vector. 

To quantize continuous local content vectors into discrete ones, we learn a dictionary of local content vectors $\mathcal{D} = \{\bm{e}_k|_{k=1}^{K}\}$ where $K$ is the number of items in the dictionary. 
Then, continuous local content vectors $\{{\bm{c}}_{i}^x|_{i=1}^N\}$ are quantized into discrete local content vectors $\{ \hat{\bm{c}}_{i}^x|_{i=1}^N\}$ by looking up its nearest neighbour in the dictionary $\mathcal{D}$. Specifically, each local content vector $\bm{c}_{i}^x$ can be replaced with its nearest item $\bm{e}_k$ in the dictionary $\mathcal{D}$ by comparing the distances between $\bm{c}_{i}^x$ and all items in $\mathcal{D}$. This process can be represented by
\begin{equation}\label{eqn:quantisation}
\begin{aligned}
\hat{\bm{c}}^{x}_{i} = \bm{e}_{k}, \text { where } k = {{\arg \min}_j }\|{{\bm{c}}}_{i}^x - \bm{e}_j\|_2.
\end{aligned}
\end{equation}
Then, the sequence of discrete local content vectors $\{\hat{\bm{c}}_{i}^x|_{i=1}^N\}$ can be reorganized into a discrete content map $\hat{\bm{C}}^x \in \mathbb{R}^{w\times h\times d_c}$ according to the order from left to right and from top to bottom. As in Section~\ref{sec:funit}, translation and reconstruction is performed based on $\hat{\bm{C}}^x$.
On the one hand, $\hat{\bm{C}}^x$ and $\bm{s}^y$ are used to produce the translated image $\tilde{\bm{y}}= F(\hat{\bm{C}}^x, \bm{s}^y)$. On the other hand,  $\hat{\bm{C}}^x$ and $\bm{s}^x$ are used to reconstruct $\tilde{\bm{x}}$.
Following~\cite{bengio2013estimating}, we adopt a straight-through gradient estimator. This estimator simply copies the gradients from the decoder $F$ to the content encoder $E_c$ and achieves back-propagation through the non-differentiable quantization procedure, which allows the disentanglement model and the dictionary of local content vector $\mathcal{D}$ to be trained in an end-to-end manner with the following loss function:
\begin{equation}\label{eqn:stage1}
\begin{aligned}
\mathcal{L}_{VQ}(G,D)= \lambda_{\mathrm{vq}} {L}_{vq}  + \mathcal{L}_{fun},
\end{aligned}
\end{equation}
in which ${L}_{vq}  = \|\operatorname{sg}[\bm{C^x}] - \hat{\bm{C^x}} \|_{2}^{2} + \|\operatorname{sg}[{\hat{\bm{C^x}}}]-\bm{C^x}\|_{2}^{2}$. $sg[\cdot]$ represents the stop-gradient operation and $\|\operatorname{sg}[\bm{C^x}] - \hat{\bm{C^x}} \|_{2}^{2}$ denotes  commitment loss \cite{van2017neural}. Another loss term $\mathcal{L}_{fun}$ is identical with Eqn.~\ref{eqn:funit} and $\lambda_{\mathrm{vq}}$ is  a hyper-parameter. 

\subsection{Style-conditioned Content Map Autoregression}\label{sec:stage2}
After the first training stage, we still require a large amount of images to provide adequate discrete content maps $\hat{\bm{C}}$. To support stochastic sampling to produce diverse and plausible discrete content maps, we model the autoregressive distribution of discrete content maps of real images. 

In particular, the process of generating a discrete content map can be formulated as a sequence-to-sequence prediction. Thus, we can learn a probabilistic model $T_c$ to predict the distribution of next local content vector based on the known local content vectors. 
Given the historic sequence $\hat{\bm{C}}_{<i}=\{\hat{\bm{c}}_j|_{j=1}^{i-1}\}$ ($\hat{\bm{C}}_{<1}=\emptyset$), the probabilistic model predicts the distribution of $\hat{\bm{c}}_i$, that is, $p(\hat{\bm{c}}_i |\hat{\bm{C}}_{<i})$. In detail, each local content vector $\bm{c}^x_i$ can be encoded into one-hot vector according to its index $k$ in dictionary $\mathcal{D}$. The probabilistic model takes in the sequence $\hat{\bm{C}}_{<i}$ and predicts the probability of the index $k$ of $\hat{\bm{c}}_i$.
The probabilistic model can be realized by many sequence-to-sequence models like PixelCNN~\cite{oord2016conditional}, LSTM~\cite{xingjian2015convolutional}, and Transformers~\cite{dosovitskiy2020image, parmar2018image,yang2020learning}. We opt for Transformer due to its ability of capture long-range dependency and superior performance in various tasks~\cite{kaiser2018fast,esser2021taming}. The technical details of Transformer are left to Supplementary due to space limitation. Overall, the probabilistic model $T_c$ tends to maximize the likelihood of discrete content maps of real images $\bm{x}$:
\begin{equation}\label{eqn:prob_autoregression}
\begin{aligned}
p(\hat{\bm{C}}^x)= \Pi_{i=1}^{N} p(\hat{\bm{c}}_{i}^x \mid \hat{\bm{C}}_{<i}^x).
\end{aligned}
\end{equation}
After training the probabilistic model, we can sample $\{\hat{\bm{c}}_i|_{i=1}^N\}$ according to the predicted probability $p(\hat{\bm{c}}_i |\hat{\bm{C}}_{<i})$ in the sequential order to construct a plausible discrete content map. 
However, the sampled discrete content map may be incompatible with the style vector and the same issue also exists for previous few-shot image translation methods like FUNIT (see Section~\ref{sec:Qualitative}). Therefore, we extend $T_c$ to a style-conditioned probabilistic model $T_{c|s}$ to generate discrete content map conditioned on a given style vector. The only difference between $T_c$ and $T_{c|s}$  is that we append a style vector to the input sequence. 
Specifically, $T_{c|s}$ takes in $\bm{s}^x$ and the historic sequence $\hat{\bm{C}}_{<i}^x$ to predict the probability of $\hat{\bm{c}}_{i}^x $. Overall, the probabilistic model $T_{c|s}$ tends to maximize the conditional likelihood of discrete content maps of real images $\bm{x}$:
\begin{equation}\label{eqn:transformer}
\begin{aligned}
p(\hat{\bm{C}}^x|\bm{s}^x)= \Pi_{i=1}^{N} p(\hat{\bm{c}}_{i}^x \mid \bm{s}^x, \hat{\bm{C}}_{<i}^x).
\end{aligned}
\end{equation}
The loss function of training $T_{c|s}$ can be written as
\begin{equation}\label{eqn:transformer_optimization}
\mathcal{L}_{T}=\mathbb{E}_{\bm{x}}[-\log p(\hat{\bm{C}}^x|\bm{s}^x)].
\end{equation}
During testing, given an image $\bm{z}$ from an unseen category $l^z$, we extract its style vector $\bm{s}^z$ from style encoder $E_s$. The probabilistic model $T_{c|s}$ functions as a content map generator, which generates amounts of discrete content maps $\{\hat{\bm{C}}^z\}$ conditioned on $\bm{s}^z$. 
These discrete content maps $\{\hat{\bm{C}}^z\}$ and $\bm{s}^z$ are fed into the decoder $F$ to produce new images $\{\tilde{\bm{z}}\}$ belonging to category $l^z$.

\section{Experiments}
\label{sec:experiments}
\subsection{Experimental Setup} 
\noindent\textbf{Datasets} We conduct experiment on Flowers~\cite{nilsback2008automated}, Animal Faces~\cite{deng2009imagenet}, and NABirds~\cite{van2015building} datasets. Following the split setting of FUNIT~\cite{liu2019few}, a total of $102$ (\emph{resp.}, $149$, and $555$) categories of Flowers (\emph{resp.}, Animal Faces, and NABirds) dataset are split into $85$ (\emph{resp.}, $119$, and $444$) seen categories and $17$ (\emph{resp.}, $30$, and $111$) unseen categories. 
\setlength{\tabcolsep}{4pt}
\begin{table*}[t]
  \caption{FID ($\downarrow$) and LPIPS ($\uparrow$) of images generated by different methods for unseen categories on three datasets.} 
  \centering
  \resizebox{1.4\columnwidth}{!} {
  \begin{tabular}{l|l|rr|rr|rr}
      \toprule[0.8pt]
      \multirow{2}{*}{Method} & 
      \multirow{2}{*}{Setting} &
      \multicolumn{2}{c|}{Flowers} &
      \multicolumn{2}{c|}{Animal Faces} & 
      \multicolumn{2}{c}{NABirds}  \cr
      & & FID ($\downarrow$)  & LPIPS ($\uparrow$) &FID ($\downarrow$) & LPIPS ($\uparrow$) & FID ($\downarrow$)  & LPIPS ($\uparrow$)   \cr
      \cmidrule(r){1-1} \cmidrule(r){2-2}
      \cmidrule(r){3-4}  \cmidrule(r){5-6}  \cmidrule(r){7-8} 
    
    DAWSON~\cite{liang2020dawson} &3-shot &  188.96 &0.0583  & 208.68  &0.0642 &181.97 &0.1105 \cr
    MatchingGAN~\cite{hong2020matchinggan} &3-shot&  143.35&0.1627 & 148.52& 0.1514& 142.52 &0.1915 \cr
    F2GAN~\cite{hong2020f2gan} &3-shot& 120.48 &0.2172 &117.74  &0.1831 &126.15 &0.2015 \cr
    
    DAGAN~\cite{antoniou2017data} &3-shot&  151.21&0.0812 &155.29 &0.0892 & 159.69 &0.1405 \cr
    DeltaGAN~\cite{hong2020deltagan} &3-shot &104.62& 0.4281& 87.04 &0.4642& 95.97&0.5136 \cr
    FUNIT~\cite{liu2019few} &3-shot &100.92  &0.4717 &86.54&0.4748&92.22&0.5689\cr
    COCO-FUNIT~\cite{saito2020coco} &3-shot &98.78 & 0.4681&83.22&0.4853&84.73&0.5617	\cr
    Disco-FUNIT (1st stage) &3-shot &$\textit{86.45}$ & $\textit{0.5009}$& $\textit{69.84}$ & $\textit{0.4928}$& $\textit{73.11}$&$\textit{0.5733}$ \cr
    Disco-FUNIT &3-shot &$\textbf{84.15}$ & $\textbf{0.5143}$& $\textbf{66.05}$ & $\textbf{0.5008}$& $\textbf{69.25}$&$\textbf{0.5842}$ \cr
    \hline
    DAGAN~\cite{antoniou2017data} &1-shot &179.59 &0.0496 & 185.54 &0.0687 & 183.57 &0.0967 \cr
    DeltaGAN~\cite{hong2020deltagan} &1-shot &109.78 &0.3912 &89.81 & 0.4418&96.79 &0.5069 \cr
    FUNIT~\cite{liu2019few} &1-shot &105.65 &0.4221 &88.07&0.4362& 	90.48& 0.4948	\cr
    COCO-FUNIT~\cite{saito2020coco} &1-shot &100.61 & 0.4197&84.95&0.4393	 &	88.55 &  0.4772\cr
    Disco-FUNIT (1st stage) &1-shot &$\textit{93.59}$ &$\textit{0.4373}$ &$\textit{75.47}$  & $\textit{0.4423}$&$\textit{79.45}$ &$\textit{0.5084}$ \cr
    Disco-FUNIT &1-shot &$\textbf{90.12}$ &$\textbf{0.4436}$ &$\textbf{71.44}$  & $\textbf{0.4511}$&$\textbf{76.12}$ &$\textbf{0.5121}$ \cr
    \bottomrule[0.8pt]
  \end{tabular}
  }
  \label{tab:performance_metric}
\end{table*}

\noindent\textbf{Implementation} Following FUNIT~\cite{liu2019few}, we set $\lambda_{\mathrm{R}}=0.1$, $\lambda_{\mathrm{F}} = 1$ without further tuning. We set $\lambda_{\mathrm{vq}}=0.8$, $K=1024$, and $N=16\times16$ after a few trials by observing the quality of generated images in the training stage. We use Pytorch $1.7.0$ to implement our model, which is distributed on RTX 2080 Ti GPU. For the first (\emph{resp.}, second) stage, we employ Adam optimizer with learning rate of $1e-4$ (\emph{resp.,} $5e-5$), and the batch size is set to $8$  (\emph{resp.,} $4$). Following~\cite{liu2019few},  image resolution in our experiments is $128\times128$. 

\noindent\textbf{Baselines} We compare our method Disco-FUNIT with few-shot image translation methods including FUNIT~\cite{liu2019few} and COCO-FUNIT~\cite{saito2020coco}, as well as few-shot image generation methods with same setting as ours including
DAWSON~\cite{liang2020dawson}, MatchingGAN~\cite{hong2020matchinggan}, F2GAN~\cite{hong2020f2gan}, DAGAN~\cite{antoniou2017data}, and DeltaGAN~\cite{hong2020deltagan}. Note that few-shot image translation methods rely on seen images at test time.
Besides, we can use our method without the second training stage as a quantized few-shot image translation method, which is referred to as ``Disco-FUNIT (1st stage)''.

\subsection{Quantitative Evaluation} \label{sec:quantitative}
We perform quantitative evaluation on the realism and diversity of the generated images, and also evaluate the category-preserving property using the downstream few-shot classification task.

\setlength{\tabcolsep}{2pt}
\begin{table*}[t]
  \caption{Accuracy(\%) of different methods on three datasets in few-shot classification setting. Note that fusion-based methods MatchingGAN~\cite{hong2020matchinggan} and F2GAN~\cite{hong2020f2gan} are not usable in $1$-shot setting.} 
  \centering
  \resizebox{1.8\columnwidth}{!} {
  \begin{tabular}{l|rr|rr|rr}
      \toprule[0.8pt]
      \multirow{2}{*}{Method}&
      \multicolumn{2}{c|}{Flowers}&
      \multicolumn{2}{c|}{Animal Faces} & 
      \multicolumn{2}{c}{NABirds}\cr
      & 10-way 1-shot &10-way 5-shot & 10-way 1-shot &10-way 5-shot & 10-way 1-shot &10-way 5-shot\cr
      \cmidrule(r){1-1} \cmidrule(r){2-2}
      \cmidrule(r){2-3}  \cmidrule(r){4-5}  \cmidrule(r){6-7}
    MatchingNets &40.96 &56.12  & 36.54 &50.12 &33.59 &46.08\cr

    MAML   &42.95 &58.01  &35.98   &49.89 & 34.12&46.21\cr

    RelationNets &48.18 &61.03  & 45.32 & 58.12 &40.59 &49.68\cr

    MTL~\cite{sun2019meta}  &54.34 &73.24  &52.54  &70.91 &44.21 &59.64\cr

    DN4~\cite{li2019revisiting}  &56.76 &73.96 & 53.26 &71.34 & 43.53&60.51\cr
    
    MatchingNet-LFT~\cite{Hungfewshot}   & 58.41 &74.32  & 56.83 &71.62 &46.16 &60.67\cr
    
    DPGN~\cite{yang2020dpgn} & 58.95&74.56&57.18 &72.02 & 46.39&58.38\cr 
    
    DeepEMD~\cite{zhang2020deepemd}    &59.12 &73.97  &58.01  &72.71& 47.68&58.29 \cr
    GCNET~\cite{Liu9343776} &57.61 &72.47 & 56.64 &71.53 &46.35 &60.36\cr
    MatchingGAN~\cite{hong2020matchinggan} &- & 74.09 & - &  70.89& - &57.16\cr

    F2GAN~\cite{hong2020f2gan}  &- &75.02  &-  &73.19 &- &59.78\cr

    DeltaGAN~\cite{hong2020deltagan} &61.23 & 77.09&60.31 & 74.59&49.05 &61.58\cr

    FUNIT~\cite{liu2019few} &55.98 & 73.12&56.61 &69.12&46.08 &56.84\cr

    COCO-FUNIT~\cite{saito2020coco} &57.18 & 74.91&58.01 &72.78 &47.58 &59.04\cr
    
    Disco-FUNIT (1st stage) &\textit{62.59} & \textit{77.52}&\textit{60.96} &\textit{75.11} & $\textit{50.34}$ & $\textit{61.98}$\cr
    Disco-FUNIT &$\textbf{63.11}$ & $\textbf{78.89}$&$\textbf{61.85}$ & $\textbf{75.71}$ & $\textbf{51.84}$ & $\textbf{63.18}$\cr
    \bottomrule[0.8pt]
    \end{tabular}
    }
  \label{tab:performance_fewshot_classifier}
\end{table*}

\begin{figure*}
\begin{center}
\includegraphics[scale=0.8]{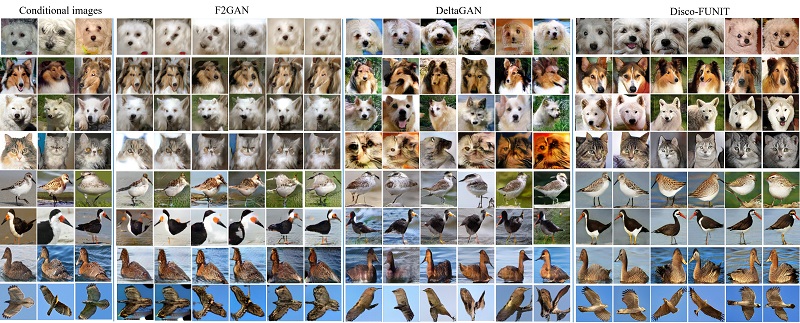}
\end{center}
\caption{Images generated by F2GAN~\cite{hong2020f2gan}, DeltaGAN~\cite{hong2020deltagan}, and our Disco-FUNIT in 3-shot setting on two datasets (from top to bottom: Animal Faces and NABirds). The conditional images are in the left three columns.}
\label{fig:visualization_compare} 
\end{figure*}


\noindent\textbf{Evaluation of Realism and Diversity}
To evaluate the realism and diversity of images generated by different methods, Fréchet Inception Distance (FID)~\cite{heusel2017gans} and Learned Perceptual Image Patch Similarity (LPIPS)~\cite{zhang2018unreasonable} are calculated on three datasets. 
We use FID to measure the distance between generated unseen images and real unseen images. 
In particular, we use ImageNet-pretrained Inception-V3~\cite{szegedy2016rethinking} model as the feature extractor. Then, FID is calculated between the extracted features of generated unseen images and those of real unseen images. 
The LPIPS is used to assess the diversity of generated unseen images. The average of pairwise distances among generated images is computed for each unseen category, and the final LPIPS score is obtained by averaging over all unseen categories. 

Considering that the number of conditional images in fusion-based methods MatchingGAN and F2GAN is configurable, we use 3 conditional images following F2GAN~\cite{hong2020f2gan}. If $H$ images are provided for each unseen category at test time, we refer to this configuration as the $H$-shot setting. We report the $3$-shot results for all methods, and also report $1$-shot results for DAGAN, DeltaGAN, FUNIT, COCO-FUNIT, and our method. Following \cite{hong2020deltagan}, we employ each method to generate $128$ images for each unseen category for calculating FID and LPIPS either in $1$-shot setting or $3$-shot setting. We also try using more images to calculate FID, but observe no significant difference.

For DAGAN, DeltaGAN, FUNIT, COCO-FUNIT, and our method in $3$-shot setting, we generate $128$ images by randomly sampling one conditional/style image each time. We summarize the results in Table \ref{tab:performance_metric}. In both  $3$-shot and $1$-shot settings, we can see that our method has the lowest FID and the highest LPIPS, indicating that it can generate more diverse and realistic images than baseline methods. Disco-FUNIT (1st stage) has already outperformed FUNIT and COCO-FUNIT, which demonstrates that discrete content maps can help remove noisy and redundant information from continuous content maps for better disentanglement. The second stage in our method can further improve the performance (Disco-FUNIT \emph{v.s.} Disco-FUNIT (1st stage)), which shows that our style-conditioned content map generator can sample diverse content maps compatible with the given style vector for producing new images of higher fidelity. Another observation is that few-shot image translation methods (\emph{i.e.}, FUNIT and COCO-FUNIT) generally outperform few-shot image generation baselines by using seen images to provide content information during testing. 

\noindent\textbf{Few-shot Classification}
To demonstrate that our generated images belong to the desired category and can benefit few-shot classification, we conduct experiments in $L$-way $M$-shot setting following~\cite{vinyals2016matching,sung2018learning}, in which the average accuracy over several evaluation episodes is calculated. In each evaluation episode, $L$ unseen categories are chosen randomly and $M$ images are chosen randomly from each selected unseen category. The remaining images from $L$ unseen categories form the test set, while the selected $L \times M$ images form the training set. We use the seen images to pretrain ResNet$18$~\cite{he2016deep} and remove the last FC layer to extract features for unseen images. In $L$-way $M$-shot setting, following~\cite{hong2020f2gan}, our method generates $512$ new images to augment each of $L$ categories in each evaluation episode. We extract features of $L\times(M+512)$ unseen images in the training set and train a linear classifier, which is then applied to the test set.

\begin{figure}
\begin{center}
\includegraphics[scale=0.15]{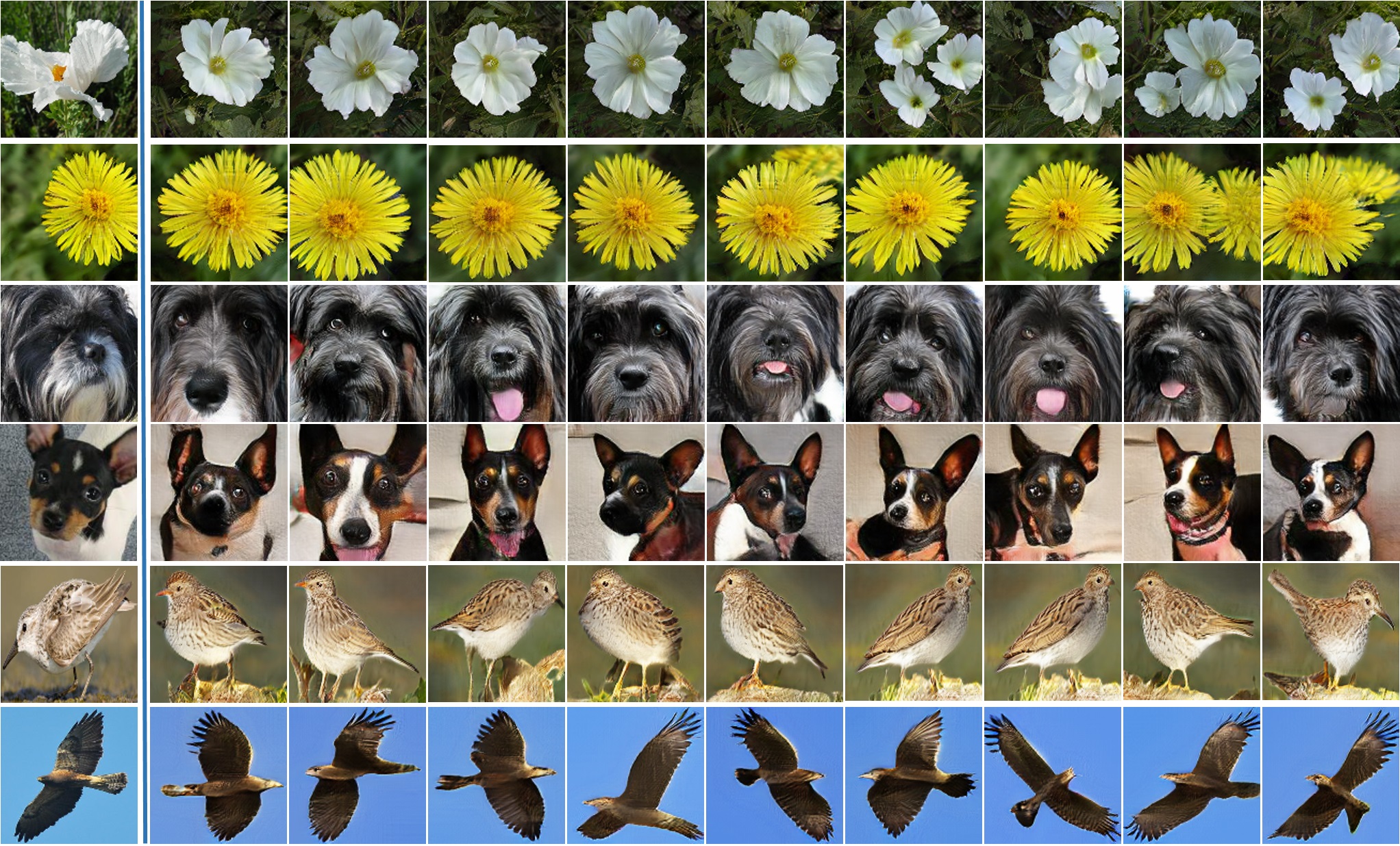}
\end{center}
\caption{Images generated by our Disco-FUNIT in 1-shot setting on three datasets (from top to bottom:  Flowers, Animal Faces, and NABirds). The conditional images are in the leftmost column.}
\label{fig:visualization} 
\end{figure}

We compare our method with state-of-the-art few-shot classification methods, including 
the representative methods MatchingNets \cite{vinyals2016matching}, RelationNets \cite{sung2018learning}, MAML \cite{finn2017model} as well as
the state-of-the-art methods 
MTL~\cite{sun2019meta}, DN4~\cite{li2019revisiting}, MatchingNet-LFT~\cite{Hungfewshot}, DPGN~\cite{yang2020dpgn}, DeepEMD~\cite{zhang2020deepemd}, and GCNET~\cite{Liu9343776}. In each evaluation episode, no augmented images are added to the training set for these baselines. Instead, images from seen categories are used to train those few-shot classifiers while adhering to their original training strategy. 
MAML \cite{finn2017model} and MTL~\cite{sun2019meta} models must be fine-tuned in each evaluation episode based on the training set. 
We also compare our method with few-shot image generation methods MatchingGAN, F2GAN, DeltaGAN as well as few-shot image translation methods FUNIT, COCO-FUNIT. We adopt the same augmentation strategy as our method in each evaluation episode. 
We show the averaged accuracy across $10$ evaluation episodes on three datasets in Table~\ref{tab:performance_fewshot_classifier} using $10$-way $1$-shot/$5$-shot as examples. It can be seen that our method produces the best results on all datasets, demonstrating the high quality of images generated by our method. Another observation is that few-shot image translation methods (\emph{i.e.}, FUNIT and COCO-FUNIT) underperform some few-shot image generation methods (\emph{e.g.}, DeltaGAN, our full method), because the content map
may also contain style information and disturb the category of generated images~\cite{hong2020deltagan}.

\subsection{Qualitative Evaluation} \label{sec:Qualitative}

\noindent\textbf{Comparison with Few-shot Image Generation Baselines}
We exhibit some example images generated by F2GAN, DeltaGAN, and our method conditioned on $3$ unseen images in Figure~\ref{fig:visualization_compare}. For DeltaGAN and our method, the generated images are arranged according to the conditional images (two generated images for each conditional image). We can observe that the images generated by F2GAN are still similar to one of the conditional images, while DeltaGAN may generate distorted images with blurry details. In contrast, our method can generate images of higher diversity and fidelity,
because our discrete content map can help remove redundant and noisy information and style-conditioned content map generator can sample content maps compatible with style vectors of the conditional images. 
In Figure~\ref{fig:visualization}, we also show some example images generated by our method on three datasets in $1$-shot setting. 
More comparison images are visualized in Supplementary. 

In Figure~\ref{fig:visualization}, we also show some example images generated by our method on three datasets in $1$-shot setting. In detail, given an unseen image, conditioned on its style vector, we can sample $9$ different style-aware discrete content maps from learned transformer, which combined with style code is fed into decoder to generate $9$ images. We can see that the generated images can keep the style of the unseen image and have desired content information (such as the pose of animals and birds, and sketch of flowers). More results can be found in Supplementary.
\begin{figure}
\begin{center}
\includegraphics[scale=0.38]{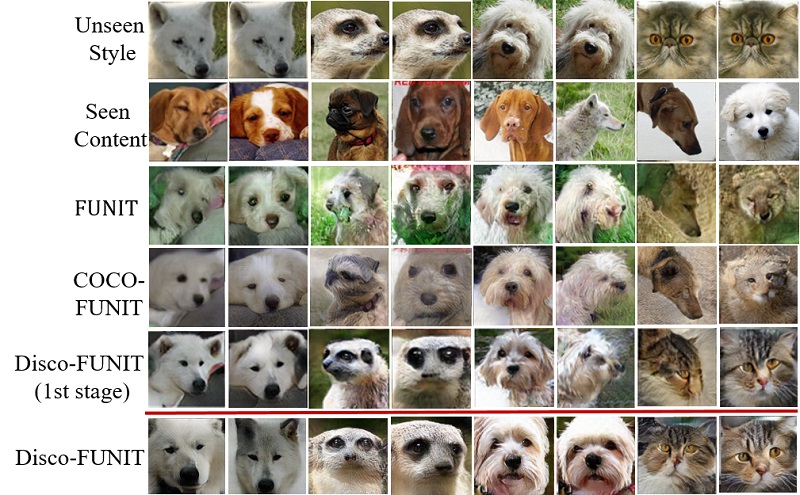}
\end{center}
\caption{Images generated by FUNIT, COCO-FUNIT, ``Disco-FUNIT (1st stage)
'', and our full Disco-FUNIT in 1-shot setting on Animal Faces dataset.}
\label{fig:process} 
\end{figure}

\noindent\textbf{Comparison with Few-shot Image Translation Baselines}
We visualize some example images generated by FUNIT, COCO-FUNIT, Disco-FUNIT (1st stage), and our full method in Figure~\ref{fig:process}. For FUNIT, COCO-FUNIT, Disco-FUNIT (1st stage), the images are generated based on the style image in the first row and the content image in the second row. For our full method, we sample content maps without using the content image in the second row.
We can observe that the images generated by FUNIT contain many artifacts, whereas COCO-FUNIT can generate higher quality images, but translated images sometimes lose details and look far from style image (see column $3$-$4$ and $7$-$8$). In contrast, images generated by Disco-FUNIT (1st stage) are more realistic and maintain the details of unseen style images (\emph{e.g.}, ear in column $1$, $7$, $8$, eye in column $3$-$4$, and texture in column $4$-$8$), since discrete content maps can remove redundant and noisy information from continuous content maps to achieve better disentanglement. However, the above methods may generate poor images (see column $4$ and $6$), when the style vector is incompatible with the content map (\emph{e.g.}, the style image and content image have too divergent poses). In contrast, our full method can sample compatible content maps to produce images with appearance details similar to the style image and plausible poses, which demonstrates the effectiveness of our style-conditioned content map generator. More generated images from Disco-FUNIT (1st stage) (\emph{resp.,} Disco-FUNIT) are visualized in Supplementary.
\setlength{\tabcolsep}{5pt}
\begin{table}[t]
  \caption{Ablation studies of each stage and alternative designs of content map generator on Animal Faces dataset.} 
  \centering
  \resizebox{0.8\columnwidth}{!} {
  \begin{tabular}{l|r|r|r}  
    \hline
     Setting&  FID $\downarrow$  & LPIPS $\uparrow$ & Accuracy(\%) $\uparrow$ \cr
    \hline
    
     FUNIT~\cite{liu2019few} & 88.07 &  0.4362  & 56.61  \cr
     
     Disco-FUNIT (1st stage) &75.47  &0.4423    &60.96   \cr

     
     \hline
     $p(\hat{\bm{C}}^x)$ & 74.43  & \textbf{0.4601}   & 60.29  \cr
     
     
     $p(\hat{\bm{C}}^x|\bm{s}^{y})$ & 78.52 & 0.4523   &59.08   \cr
     


    
    
    
    \hline
    Disco-FUNIT & \textbf{71.44} & 0.4511  & \textbf{61.85}  \cr
    \hline
  \end{tabular}
  }
  \label{tab:ablation_study}
\end{table}

    
     

     
     
     
     


    
    
    

\subsection{Ablation Studies}
\label{sec:ablation_study}
By taking Animal Faces dataset as an example, we study the effect of each stage in our method and alternative network designs. 
For each ablated method, we report FID, LPIPS, and the accuracy of few-shot classification (10-way 1-shot) as in Section~\ref{sec:quantitative} in Table~\ref{tab:ablation_study}. 
The analyses of hyper-parameters (\emph{e.g.}, content map size, dictionary length, and transformer layers) are left to Supplementary.
due to space limitation. 

\noindent\textbf{Effect of Each Stage in Our Method}
First, we remove the second training stage from our method, which is referred to as  ``Disco-FUNIT (1st stage)'' in Table~\ref{tab:ablation_study}. The difference between FUNIT and ``Disco-FUNIT (1st stage)'' is that content map is continuous in FUNIT but discrete in ``Disco-FUNIT (1st stage)''. Based on Table~\ref{tab:ablation_study} and Figure~\ref{fig:process}, we can observe that discrete content map can substantially improve the quality of translated images, because content map quantisation can cooperate with FUNIT to remove redundant information from content map for better disentanglement. 

To study the effect of style-conditioned content map autoregression stage, we compare  ``Disco-FUNIT (1st stage)'' with our full method in Table~\ref{tab:ablation_study}, which shows that our content map generator $T_{c|s}$ can further improve the quality of generated images by sampling content maps compatible with the style vector. Besides, our content map generator $T_{c|s}$ only relies on unseen images and learned dictionary at test time, which adapts ``Disco-FUNIT (1st stage)'' to few-shot image generation. We compare storage consumption of seen images and learned dictionary in Supplementary.

\noindent\textbf{Alternative Designs of Content Map Generator}
To evaluate the effectiveness of the style-conditioned content map generator $T_{c|s}$, we explore different ways to model the autoregressive distribution of discrete content map. The first one is ``$p(\hat{\bm{C}}^x)$'' as in Eqn.~\ref{eqn:prob_autoregression}. The second one is replacing conditional style vector $\bm{s}^x$ with the style vector from a different image $\bm{s}^{y}$,
which is referred to as ``$p(\hat{\bm{C}}^x|\bm{s}^{y})$''. 
From Table~\ref{tab:ablation_study}, we can see that ``$p(\hat{\bm{C}}^x)$'' is worse than Disco-FUNIT, which demonstrates the advantage of sampling content map conditioned on the style vector. We also observe that  ``$p(\hat{\bm{C}}^x|\bm{s}^{y})$'' is even worse than ``Disco-FUNIT (1st stage)'', which proves that it is necessary to model the conditional likelihood based on paired contents map and style vector from real images.

\section{Investigation of Style-conditioned Content Map Autoregression}\label{sec:study_content_map_autoregression}
To further investigate the effectiveness of our style-conditioned content map generator $T_{c|s}$ in the second stage, we show some example images generated by Disco-FUNIT (1st stage), one variant of Disco-FUNIT (``$p(\hat{\bm{C}}^x)$''), and Disco-FUNIT. As introduced in Section~\ref{sec:ablation_study}, $p(\hat{\bm{C}}^x)$ models the autoregressive distribution of content maps without the conditional style vector.

\begin{figure}
\begin{center}
\includegraphics[scale=0.13]{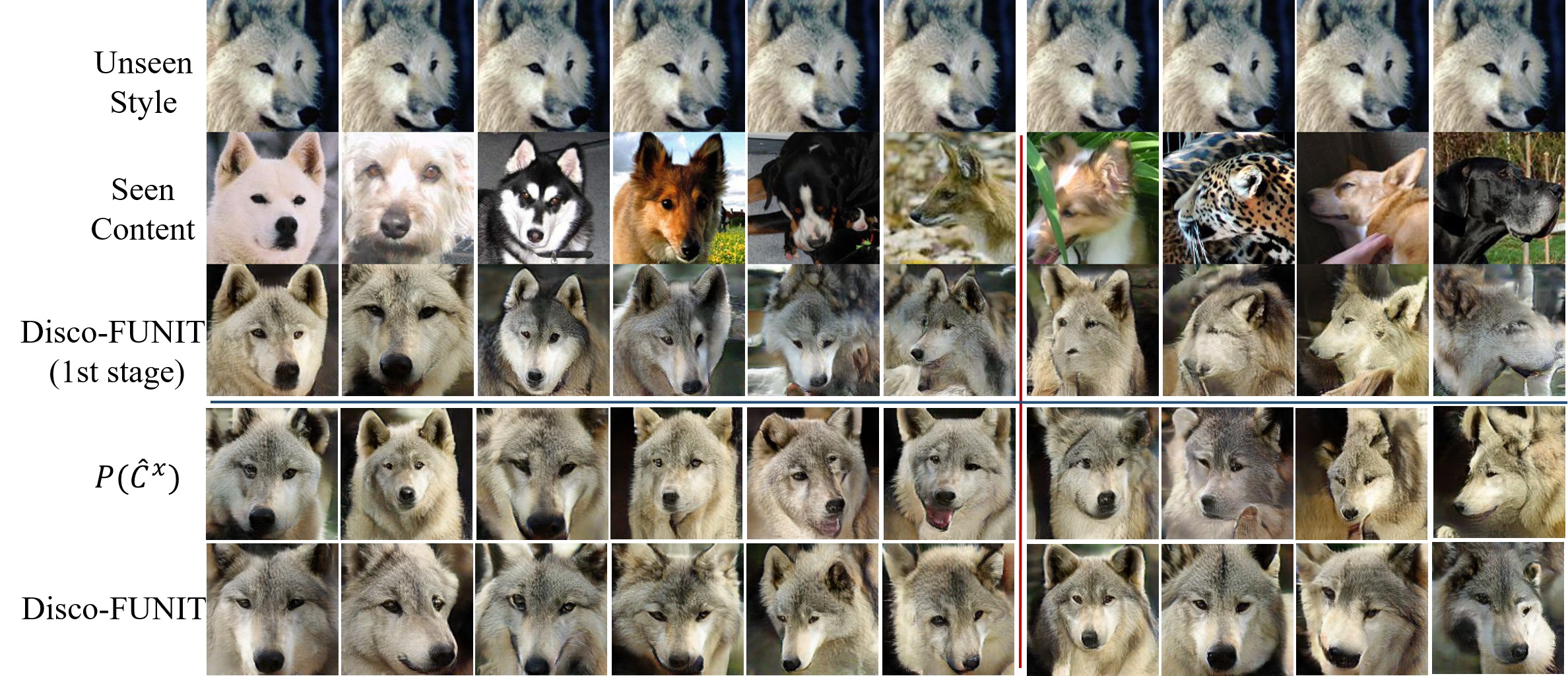}
\end{center}
\caption{Images generated by Disco-FUNIT (1st stage), ablated method ``$p(\hat{\bm{C}}^x)$'',  and Disco-FUNIT on Animal Faces dataset. (From top to bottom: unseen style images, seen content images, images generated by Disco-FUNIT (1st stage), our variant ``$p(\hat{\bm{C}}^x)$'', and Disco-FUNIT. The results of ``$p(\hat{\bm{C}}^x)$'' and Disco-FUNIT are generated by sampling content maps without using content images in row 2. The generated images in each row are arranged in a decreasing order based on the compatibility between content map and style vector. }
\label{fig:compatibility_12stage} 
\end{figure}

For Disco-FUNIT (1st stage), the images are generated based on the unseen style image in the first row and the seen content image in the second row.
In detail, given a style image, we randomly select $50$ seen content images to produce $50$ translated images. Then, we calculate the loss value of each translated image according to Eqn~\ref{eqn:transformer_optimization}, in which the smaller loss value indicates that the discrete content map and the style vector are more compatible. 
After sorting these $50$ images by loss value in an increasing order, we show six images with the smallest loss values and four images with the largest loss values in row 3 in Figure~\ref{fig:compatibility_12stage}.

For ``$p(\hat{\bm{C}}^x)$'' and Disco-FUNIT, we sample content maps without using the seen content image in the second row. Similar to Disco-FUNIT (1st stage), we sampling $50$ discrete content maps to produce $50$ images and calculate their loss values. 
Then, we show six images with the smallest loss values and four images with the largest loss values for ``$p(\hat{\bm{C}}^x)$'' (\emph{resp.,} Disco-FUNIT) in row 4 (\emph{resp.,} 5) in Figure~\ref{fig:compatibility_12stage}.

By comparing Disco-FUNIT (1st stage) and Disco-FUNIT, it can be seen that our content map generator can sample diverse content maps to produce more satisfactory images with high probability, while Disco-FUNIT (1st stage) may produce undesired images due to incompatible content maps of seen content images (see column 5-10 in row 3). By comparing ``$p(\hat{\bm{C}}^x)$'' and Disco-FUNIT, we can see that the images generated by ``$p(\hat{\bm{C}}^x)$'' are diverse but occasionally distorted, \emph{e.g.}, partial images with small loss values (column 5-6 in row 4) and all images with large loss values (column 7-10 in row 4). The observation is consistent with higher FID and higher LPIPS reported in Table ~\ref{tab:ablation_study}, which also indicates that the generated images with rich diversity may be unrealistic due to distortion or artifacts. In contrast, the images generated by Disco-FUNIT with small loss values are rich in details and look more realistic. However, we acknowledge that the images generated by Disco-FUNIT with large loss values may also be distorted (column 9-10 in row 5).


\section{Conclusion}
In this paper, we have proposed to adapt few-shot image translation method to few-shot image generation by learning a compact dictionary of local content vectors and model the compatibility between content map and style vector. Extensive experiments on three real datasets have demonstrated the effectiveness of our proposed Disco-FUNIT. 


\begin{acks}
The work is supported by Shanghai Municipal Science and Technology Major Project, China (2021SHZDZX0102), and Shanghai Municipal Science and Technology Key Project (Grant No. 20511100300), and National Science Foundation of China (61902247).
\end{acks}

\bibliographystyle{ACM-Reference-Format}
\balance
\bibliography{egbib}


\begin{thebibliography}{52}


\ifx \showCODEN    \undefined \def \showCODEN     #1{\unskip}     \fi
\ifx \showDOI      \undefined \def \showDOI       #1{#1}\fi
\ifx \showISBNx    \undefined \def \showISBNx     #1{\unskip}     \fi
\ifx \showISBNxiii \undefined \def \showISBNxiii  #1{\unskip}     \fi
\ifx \showISSN     \undefined \def \showISSN      #1{\unskip}     \fi
\ifx \showLCCN     \undefined \def \showLCCN      #1{\unskip}     \fi
\ifx \shownote     \undefined \def \shownote      #1{#1}          \fi
\ifx \showarticletitle \undefined \def \showarticletitle #1{#1}   \fi
\ifx \showURL      \undefined \def \showURL       {\relax}        \fi
\providecommand\bibfield[2]{#2}
\providecommand\bibinfo[2]{#2}
\providecommand\natexlab[1]{#1}
\providecommand\showeprint[2][]{arXiv:#2}

\bibitem[Antoniou et~al\mbox{.}(2017)]%
        {antoniou2017data}
\bibfield{author}{\bibinfo{person}{Antreas Antoniou}, \bibinfo{person}{Amos
  Storkey}, {and} \bibinfo{person}{Harrison Edwards}.}
  \bibinfo{year}{2017}\natexlab{}.
\newblock \showarticletitle{Data augmentation generative adversarial networks}.
\newblock \bibinfo{journal}{\emph{arXiv preprint arXiv:1711.04340}}
  (\bibinfo{year}{2017}).
\newblock


\bibitem[Bartunov and Vetrov(2018)]%
        {bartunov2018few}
\bibfield{author}{\bibinfo{person}{Sergey Bartunov} {and}
  \bibinfo{person}{Dmitry Vetrov}.} \bibinfo{year}{2018}\natexlab{}.
\newblock \showarticletitle{Few-shot generative modelling with generative
  matching networks}. In \bibinfo{booktitle}{\emph{AISTATS}}.
\newblock


\bibitem[Benaim and Wolf(2018)]%
        {benaim2018one}
\bibfield{author}{\bibinfo{person}{Sagie Benaim} {and} \bibinfo{person}{Lior
  Wolf}.} \bibinfo{year}{2018}\natexlab{}.
\newblock \showarticletitle{One-Shot Unsupervised Cross Domain Translation}. In
  \bibinfo{booktitle}{\emph{NeurIPS}}.
\newblock


\bibitem[Bengio et~al\mbox{.}(2013)]%
        {bengio2013estimating}
\bibfield{author}{\bibinfo{person}{Yoshua Bengio}, \bibinfo{person}{Nicholas
  L{\'e}onard}, {and} \bibinfo{person}{Aaron Courville}.}
  \bibinfo{year}{2013}\natexlab{}.
\newblock \showarticletitle{Estimating or propagating gradients through
  stochastic neurons for conditional computation}.
\newblock \bibinfo{journal}{\emph{arXiv preprint arXiv:1308.3432}}
  (\bibinfo{year}{2013}).
\newblock


\bibitem[Brock et~al\mbox{.}(2018)]%
        {brock2018large}
\bibfield{author}{\bibinfo{person}{Andrew Brock}, \bibinfo{person}{Jeff
  Donahue}, {and} \bibinfo{person}{Karen Simonyan}.}
  \bibinfo{year}{2018}\natexlab{}.
\newblock \showarticletitle{Large Scale GAN Training for High Fidelity Natural
  Image Synthesis}. In \bibinfo{booktitle}{\emph{ICLR}}.
\newblock


\bibitem[Clou{\^a}tre and Demers(2019)]%
        {clouatre2019figr}
\bibfield{author}{\bibinfo{person}{Louis Clou{\^a}tre} {and}
  \bibinfo{person}{Marc Demers}.} \bibinfo{year}{2019}\natexlab{}.
\newblock \showarticletitle{FIGR: Few-shot image generation with reptile}.
\newblock \bibinfo{journal}{\emph{arXiv preprint arXiv:1901.02199}}
  (\bibinfo{year}{2019}).
\newblock


\bibitem[Deng et~al\mbox{.}(2009)]%
        {deng2009imagenet}
\bibfield{author}{\bibinfo{person}{Jia Deng}, \bibinfo{person}{Wei Dong},
  \bibinfo{person}{Richard Socher}, \bibinfo{person}{Li-Jia Li},
  \bibinfo{person}{Kai Li}, {and} \bibinfo{person}{Li Fei-Fei}.}
  \bibinfo{year}{2009}\natexlab{}.
\newblock \showarticletitle{Imagenet: A large-scale hierarchical image
  database}. In \bibinfo{booktitle}{\emph{CVPR}}.
\newblock


\bibitem[Dosovitskiy et~al\mbox{.}(2020)]%
        {dosovitskiy2020image}
\bibfield{author}{\bibinfo{person}{Alexey Dosovitskiy}, \bibinfo{person}{Lucas
  Beyer}, \bibinfo{person}{Alexander Kolesnikov}, \bibinfo{person}{Dirk
  Weissenborn}, \bibinfo{person}{Xiaohua Zhai}, \bibinfo{person}{Thomas
  Unterthiner}, \bibinfo{person}{Mostafa Dehghani}, \bibinfo{person}{Matthias
  Minderer}, \bibinfo{person}{Georg Heigold}, \bibinfo{person}{Sylvain Gelly},
  {et~al\mbox{.}}} \bibinfo{year}{2020}\natexlab{}.
\newblock \showarticletitle{An Image is Worth 16x16 Words: Transformers for
  Image Recognition at Scale}. In \bibinfo{booktitle}{\emph{ICLR}}.
\newblock


\bibitem[Esser et~al\mbox{.}(2021)]%
        {esser2021taming}
\bibfield{author}{\bibinfo{person}{Patrick Esser}, \bibinfo{person}{Robin
  Rombach}, {and} \bibinfo{person}{Bjorn Ommer}.}
  \bibinfo{year}{2021}\natexlab{}.
\newblock \showarticletitle{Taming transformers for high-resolution image
  synthesis}. In \bibinfo{booktitle}{\emph{CVPR}}.
\newblock


\bibitem[Fan et~al\mbox{.}(2020)]%
        {fan2020scale}
\bibfield{author}{\bibinfo{person}{Yuchen Fan}, \bibinfo{person}{Jiahui Yu},
  \bibinfo{person}{Ding Liu}, {and} \bibinfo{person}{Thomas~S Huang}.}
  \bibinfo{year}{2020}\natexlab{}.
\newblock \showarticletitle{Scale-wise convolution for image restoration}. In
  \bibinfo{booktitle}{\emph{AAAI}}.
\newblock


\bibitem[Finn et~al\mbox{.}(2017)]%
        {finn2017model}
\bibfield{author}{\bibinfo{person}{Chelsea Finn}, \bibinfo{person}{Pieter
  Abbeel}, {and} \bibinfo{person}{Sergey Levine}.}
  \bibinfo{year}{2017}\natexlab{}.
\newblock \showarticletitle{Model-agnostic meta-learning for fast adaptation of
  deep networks}. In \bibinfo{booktitle}{\emph{ICML}}.
\newblock


\bibitem[He et~al\mbox{.}(2016)]%
        {he2016deep}
\bibfield{author}{\bibinfo{person}{Kaiming He}, \bibinfo{person}{Xiangyu
  Zhang}, \bibinfo{person}{Shaoqing Ren}, {and} \bibinfo{person}{Jian Sun}.}
  \bibinfo{year}{2016}\natexlab{}.
\newblock \showarticletitle{Deep residual learning for image recognition}. In
  \bibinfo{booktitle}{\emph{CVPR}}.
\newblock


\bibitem[Heusel et~al\mbox{.}(2017)]%
        {heusel2017gans}
\bibfield{author}{\bibinfo{person}{Martin Heusel}, \bibinfo{person}{Hubert
  Ramsauer}, \bibinfo{person}{Thomas Unterthiner}, \bibinfo{person}{Bernhard
  Nessler}, {and} \bibinfo{person}{Sepp Hochreiter}.}
  \bibinfo{year}{2017}\natexlab{}.
\newblock \showarticletitle{{GAN}s trained by a two time-scale update rule
  converge to a local nash equilibrium}. In
  \bibinfo{booktitle}{\emph{NeurIPS}}.
\newblock


\bibitem[Hong et~al\mbox{.}(2020a)]%
        {hong2020matchinggan}
\bibfield{author}{\bibinfo{person}{Yan Hong}, \bibinfo{person}{Li Niu},
  \bibinfo{person}{Jianfu Zhang}, {and} \bibinfo{person}{Liqing Zhang}.}
  \bibinfo{year}{2020}\natexlab{a}.
\newblock \showarticletitle{Matchinggan: Matching-Based Few-Shot Image
  Generation}. In \bibinfo{booktitle}{\emph{ICME}}.
\newblock


\bibitem[Hong et~al\mbox{.}(2022)]%
        {hong2020deltagan}
\bibfield{author}{\bibinfo{person}{Yan Hong}, \bibinfo{person}{Li Niu},
  \bibinfo{person}{Jianfu Zhang}, {and} \bibinfo{person}{Liqing Zhang}.}
  \bibinfo{year}{2022}\natexlab{}.
\newblock \showarticletitle{Delta{GAN}: Towards diverse few-shot image
  generation with sample-specific delta}.
\newblock \bibinfo{journal}{\emph{ECCV}} (\bibinfo{year}{2022}).
\newblock


\bibitem[Hong et~al\mbox{.}(2020b)]%
        {hong2020f2gan}
\bibfield{author}{\bibinfo{person}{Yan Hong}, \bibinfo{person}{Li Niu},
  \bibinfo{person}{Jianfu Zhang}, \bibinfo{person}{Weijie Zhao},
  \bibinfo{person}{Chen Fu}, {and} \bibinfo{person}{Liqing Zhang}.}
  \bibinfo{year}{2020}\natexlab{b}.
\newblock \showarticletitle{F2GAN: Fusing-and-Filling GAN for Few-shot Image
  Generation}. In \bibinfo{booktitle}{\emph{ACM MM}}.
\newblock


\bibitem[Hu et~al\mbox{.}(2020)]%
        {hu2020coarse}
\bibfield{author}{\bibinfo{person}{Yueyu Hu}, \bibinfo{person}{Wenhan Yang},
  {and} \bibinfo{person}{Jiaying Liu}.} \bibinfo{year}{2020}\natexlab{}.
\newblock \showarticletitle{Coarse-to-fine hyper-prior modeling for learned
  image compression}. In \bibinfo{booktitle}{\emph{AAAI}}.
\newblock


\bibitem[Kaiser et~al\mbox{.}(2018)]%
        {kaiser2018fast}
\bibfield{author}{\bibinfo{person}{Lukasz Kaiser}, \bibinfo{person}{Samy
  Bengio}, \bibinfo{person}{Aurko Roy}, \bibinfo{person}{Ashish Vaswani},
  \bibinfo{person}{Niki Parmar}, \bibinfo{person}{Jakob Uszkoreit}, {and}
  \bibinfo{person}{Noam Shazeer}.} \bibinfo{year}{2018}\natexlab{}.
\newblock \showarticletitle{Fast decoding in sequence models using discrete
  latent variables}. In \bibinfo{booktitle}{\emph{ICML}}.
\newblock


\bibitem[Karras et~al\mbox{.}(2019)]%
        {stylegan1}
\bibfield{author}{\bibinfo{person}{Tero Karras}, \bibinfo{person}{Samuli
  Laine}, {and} \bibinfo{person}{Timo Aila}.} \bibinfo{year}{2019}\natexlab{}.
\newblock \showarticletitle{A style-based generator architecture for generative
  adversarial networks}. In \bibinfo{booktitle}{\emph{CVPR}}.
\newblock


\bibitem[Karras et~al\mbox{.}(2020)]%
        {stylegan2}
\bibfield{author}{\bibinfo{person}{Tero Karras}, \bibinfo{person}{Samuli
  Laine}, \bibinfo{person}{Miika Aittala}, \bibinfo{person}{Janne Hellsten},
  \bibinfo{person}{Jaakko Lehtinen}, {and} \bibinfo{person}{Timo Aila}.}
  \bibinfo{year}{2020}\natexlab{}.
\newblock \showarticletitle{Analyzing and improving the image quality of
  styleGAN}. In \bibinfo{booktitle}{\emph{CVPR}}.
\newblock


\bibitem[Kolesnikov and Lampert(2017)]%
        {kolesnikov2017pixelcnn}
\bibfield{author}{\bibinfo{person}{Alexander Kolesnikov} {and}
  \bibinfo{person}{Christoph~H Lampert}.} \bibinfo{year}{2017}\natexlab{}.
\newblock \showarticletitle{PixelCNN models with auxiliary variables for
  natural image modeling}. In \bibinfo{booktitle}{\emph{ICML}}.
\newblock


\bibitem[Lake et~al\mbox{.}(2011)]%
        {lake2011one}
\bibfield{author}{\bibinfo{person}{Brenden~M Lake}, \bibinfo{person}{Ruslan
  Salakhutdinov}, \bibinfo{person}{Jason Gross}, {and}
  \bibinfo{person}{Joshua~B Tenenbaum}.} \bibinfo{year}{2011}\natexlab{}.
\newblock \showarticletitle{One shot learning of simple visual concepts}.
\newblock \bibinfo{journal}{\emph{Cognitive Science}} \bibinfo{volume}{33},
  \bibinfo{number}{33} (\bibinfo{year}{2011}).
\newblock


\bibitem[Li et~al\mbox{.}(2019)]%
        {li2019revisiting}
\bibfield{author}{\bibinfo{person}{Wenbin Li}, \bibinfo{person}{Lei Wang},
  \bibinfo{person}{Jinglin Xu}, \bibinfo{person}{Jing Huo},
  \bibinfo{person}{Yang Gao}, {and} \bibinfo{person}{Jiebo Luo}.}
  \bibinfo{year}{2019}\natexlab{}.
\newblock \showarticletitle{Revisiting local descriptor based image-to-class
  measure for few-shot learning}. In \bibinfo{booktitle}{\emph{CVPR}}.
\newblock


\bibitem[Li et~al\mbox{.}(2020)]%
        {li2020few}
\bibfield{author}{\bibinfo{person}{Yijun Li}, \bibinfo{person}{Richard Zhang},
  \bibinfo{person}{Jingwan Lu}, {and} \bibinfo{person}{Eli Shechtman}.}
  \bibinfo{year}{2020}\natexlab{}.
\newblock \showarticletitle{Few-shot image generation with elastic weight
  consolidation}. In \bibinfo{booktitle}{\emph{NeurIPS}}.
\newblock


\bibitem[Liang et~al\mbox{.}(2020)]%
        {liang2020dawson}
\bibfield{author}{\bibinfo{person}{Weixin Liang}, \bibinfo{person}{Zixuan Liu},
  {and} \bibinfo{person}{Can Liu}.} \bibinfo{year}{2020}\natexlab{}.
\newblock \showarticletitle{DAWSON: A domain adaptive few shot generation
  framework}.
\newblock \bibinfo{journal}{\emph{arXiv preprint arXiv:2001.00576}}
  (\bibinfo{year}{2020}).
\newblock


\bibitem[Liu et~al\mbox{.}(2021)]%
        {Liu9343776}
\bibfield{author}{\bibinfo{person}{Liyang Liu}, \bibinfo{person}{Bochao Wang},
  \bibinfo{person}{Zhanghui Kuang}, \bibinfo{person}{Jing-Hao Xue},
  \bibinfo{person}{Yimin Chen}, \bibinfo{person}{Wenming Yang},
  \bibinfo{person}{Qingmin Liao}, {and} \bibinfo{person}{Wayne Zhang}.}
  \bibinfo{year}{2021}\natexlab{}.
\newblock \showarticletitle{GenDet: Meta Learning to Generate Detectors From
  Few Shots}.
\newblock \bibinfo{journal}{\emph{Transactions on Neural Networks and Learning
  Systems}} (\bibinfo{year}{2021}).
\newblock


\bibitem[Liu et~al\mbox{.}(2019)]%
        {liu2019few}
\bibfield{author}{\bibinfo{person}{Ming{-}Yu Liu}, \bibinfo{person}{Xun Huang},
  \bibinfo{person}{Arun Mallya}, \bibinfo{person}{Tero Karras},
  \bibinfo{person}{Timo Aila}, \bibinfo{person}{Jaakko Lehtinen}, {and}
  \bibinfo{person}{Jan Kautz}.} \bibinfo{year}{2019}\natexlab{}.
\newblock \showarticletitle{Few-Shot Unsupervised Image-to-Image Translation}.
  In \bibinfo{booktitle}{\emph{ICCV}}.
\newblock


\bibitem[Nilsback and Zisserman(2008)]%
        {nilsback2008automated}
\bibfield{author}{\bibinfo{person}{Maria-Elena Nilsback} {and}
  \bibinfo{person}{Andrew Zisserman}.} \bibinfo{year}{2008}\natexlab{}.
\newblock \showarticletitle{Automated flower classification over a large number
  of classes}. In \bibinfo{booktitle}{\emph{CVGIP}}.
\newblock


\bibitem[Ojha et~al\mbox{.}(2021)]%
        {ojha2021few}
\bibfield{author}{\bibinfo{person}{Utkarsh Ojha}, \bibinfo{person}{Yijun Li},
  \bibinfo{person}{Jingwan Lu}, \bibinfo{person}{Alexei~A Efros},
  \bibinfo{person}{Yong~Jae Lee}, \bibinfo{person}{Eli Shechtman}, {and}
  \bibinfo{person}{Richard Zhang}.} \bibinfo{year}{2021}\natexlab{}.
\newblock \showarticletitle{Few-shot Image Generation via Cross-domain
  Correspondence}. In \bibinfo{booktitle}{\emph{CVPR}}.
\newblock


\bibitem[Oord et~al\mbox{.}(2016)]%
        {oord2016conditional}
\bibfield{author}{\bibinfo{person}{A{\"a}ron van~den Oord},
  \bibinfo{person}{Nal Kalchbrenner}, \bibinfo{person}{Oriol Vinyals},
  \bibinfo{person}{Lasse Espeholt}, \bibinfo{person}{Alex Graves}, {and}
  \bibinfo{person}{Koray Kavukcuoglu}.} \bibinfo{year}{2016}\natexlab{}.
\newblock \showarticletitle{Conditional image generation with PixelCNN
  decoders}. In \bibinfo{booktitle}{\emph{NeurIPS}}.
\newblock


\bibitem[Parmar et~al\mbox{.}(2018)]%
        {parmar2018image}
\bibfield{author}{\bibinfo{person}{Niki Parmar}, \bibinfo{person}{Ashish
  Vaswani}, \bibinfo{person}{Jakob Uszkoreit}, \bibinfo{person}{Lukasz Kaiser},
  \bibinfo{person}{Noam Shazeer}, \bibinfo{person}{Alexander Ku}, {and}
  \bibinfo{person}{Dustin Tran}.} \bibinfo{year}{2018}\natexlab{}.
\newblock \showarticletitle{Image transformer}. In
  \bibinfo{booktitle}{\emph{ICML}}.
\newblock


\bibitem[Razavi et~al\mbox{.}(2019)]%
        {razavi2019generating}
\bibfield{author}{\bibinfo{person}{Ali Razavi}, \bibinfo{person}{Aaron van~den
  Oord}, {and} \bibinfo{person}{Oriol Vinyals}.}
  \bibinfo{year}{2019}\natexlab{}.
\newblock \showarticletitle{Generating diverse high-fidelity images with
  vq-vae-2}. In \bibinfo{booktitle}{\emph{NeurIPS}}.
\newblock


\bibitem[Rezende et~al\mbox{.}(2016)]%
        {rezende2016one-shot}
\bibfield{author}{\bibinfo{person}{Danilo~Jimenez Rezende},
  \bibinfo{person}{Shakir Mohamed}, \bibinfo{person}{Ivo Danihelka},
  \bibinfo{person}{Karol Gregor}, {and} \bibinfo{person}{Daan Wierstra}.}
  \bibinfo{year}{2016}\natexlab{}.
\newblock \showarticletitle{One-shot generalization in deep generative models}.
  In \bibinfo{booktitle}{\emph{ICML}}.
\newblock


\bibitem[Robb et~al\mbox{.}(2020)]%
        {robb2020few}
\bibfield{author}{\bibinfo{person}{Esther Robb}, \bibinfo{person}{Wen-Sheng
  Chu}, \bibinfo{person}{Abhishek Kumar}, {and} \bibinfo{person}{Jia-Bin
  Huang}.} \bibinfo{year}{2020}\natexlab{}.
\newblock \showarticletitle{Few-shot adaptation of generative adversarial
  networks}.
\newblock \bibinfo{journal}{\emph{arXiv preprint arXiv:2010.11943}}
  (\bibinfo{year}{2020}).
\newblock


\bibitem[Saito et~al\mbox{.}(2020)]%
        {saito2020coco}
\bibfield{author}{\bibinfo{person}{Kuniaki Saito}, \bibinfo{person}{Kate
  Saenko}, {and} \bibinfo{person}{Ming-Yu Liu}.}
  \bibinfo{year}{2020}\natexlab{}.
\newblock \showarticletitle{Coco-funit: Few-shot unsupervised image translation
  with a content conditioned style encoder}. In
  \bibinfo{booktitle}{\emph{ECCV}}.
\newblock


\bibitem[Sun et~al\mbox{.}(2019)]%
        {sun2019meta}
\bibfield{author}{\bibinfo{person}{Qianru Sun}, \bibinfo{person}{Yaoyao Liu},
  \bibinfo{person}{Tat-Seng Chua}, {and} \bibinfo{person}{Bernt Schiele}.}
  \bibinfo{year}{2019}\natexlab{}.
\newblock \showarticletitle{Meta-transfer learning for few-shot learning}. In
  \bibinfo{booktitle}{\emph{CVPR}}.
\newblock


\bibitem[Sung et~al\mbox{.}(2018)]%
        {sung2018learning}
\bibfield{author}{\bibinfo{person}{Flood Sung}, \bibinfo{person}{Yongxin Yang},
  \bibinfo{person}{Li Zhang}, \bibinfo{person}{Tao Xiang},
  \bibinfo{person}{Philip~HS Torr}, {and} \bibinfo{person}{Timothy~M
  Hospedales}.} \bibinfo{year}{2018}\natexlab{}.
\newblock \showarticletitle{Learning to compare: Relation network for few-shot
  learning}. In \bibinfo{booktitle}{\emph{CVPR}}.
\newblock


\bibitem[Sutskever et~al\mbox{.}(2014)]%
        {sutskever2014sequence}
\bibfield{author}{\bibinfo{person}{Ilya Sutskever}, \bibinfo{person}{Oriol
  Vinyals}, {and} \bibinfo{person}{Quoc~V Le}.}
  \bibinfo{year}{2014}\natexlab{}.
\newblock \showarticletitle{Sequence to sequence learning with neural
  networks}. In \bibinfo{booktitle}{\emph{NeurIPS}}.
\newblock


\bibitem[Szegedy et~al\mbox{.}(2016)]%
        {szegedy2016rethinking}
\bibfield{author}{\bibinfo{person}{Christian Szegedy}, \bibinfo{person}{Vincent
  Vanhoucke}, \bibinfo{person}{Sergey Ioffe}, \bibinfo{person}{Jon Shlens},
  {and} \bibinfo{person}{Zbigniew Wojna}.} \bibinfo{year}{2016}\natexlab{}.
\newblock \showarticletitle{Rethinking the inception architecture for computer
  vision}. In \bibinfo{booktitle}{\emph{CVPR}}.
\newblock


\bibitem[Tseng et~al\mbox{.}(2020)]%
        {Hungfewshot}
\bibfield{author}{\bibinfo{person}{Hung{-}Yu Tseng},
  \bibinfo{person}{Hsin{-}Ying Lee}, \bibinfo{person}{Jia{-}Bin Huang}, {and}
  \bibinfo{person}{Ming{-}Hsuan Yang}.} \bibinfo{year}{2020}\natexlab{}.
\newblock \showarticletitle{Cross-domain few-shot classification via Learned
  Feature-Wise Transformation}. In \bibinfo{booktitle}{\emph{ICLR}}.
\newblock


\bibitem[van~den Oord et~al\mbox{.}(2017)]%
        {van2017neural}
\bibfield{author}{\bibinfo{person}{Aaron van~den Oord}, \bibinfo{person}{Oriol
  Vinyals}, {and} \bibinfo{person}{Koray Kavukcuoglu}.}
  \bibinfo{year}{2017}\natexlab{}.
\newblock \showarticletitle{Neural discrete representation learning}. In
  \bibinfo{booktitle}{\emph{NeurIPS}}.
\newblock


\bibitem[Van~Horn et~al\mbox{.}(2015)]%
        {van2015building}
\bibfield{author}{\bibinfo{person}{Grant Van~Horn}, \bibinfo{person}{Steve
  Branson}, \bibinfo{person}{Ryan Farrell}, \bibinfo{person}{Scott Haber},
  \bibinfo{person}{Jessie Barry}, \bibinfo{person}{Panos Ipeirotis},
  \bibinfo{person}{Pietro Perona}, {and} \bibinfo{person}{Serge Belongie}.}
  \bibinfo{year}{2015}\natexlab{}.
\newblock \showarticletitle{Building a bird recognition app and large scale
  dataset with citizen scientists: The fine print in fine-grained dataset
  collection}. In \bibinfo{booktitle}{\emph{CVPR}}.
\newblock


\bibitem[Vinyals et~al\mbox{.}(2016)]%
        {vinyals2016matching}
\bibfield{author}{\bibinfo{person}{Oriol Vinyals}, \bibinfo{person}{Charles
  Blundell}, \bibinfo{person}{Timothy Lillicrap}, \bibinfo{person}{Wierstra},
  {and} \bibinfo{person}{Daan}.} \bibinfo{year}{2016}\natexlab{}.
\newblock \showarticletitle{Matching networks for one shot learning}. In
  \bibinfo{booktitle}{\emph{NeurIPS}}.
\newblock


\bibitem[Wallace(1992)]%
        {wallace1992jpeg}
\bibfield{author}{\bibinfo{person}{Gregory~K Wallace}.}
  \bibinfo{year}{1992}\natexlab{}.
\newblock \showarticletitle{The JPEG still picture compression standard}.
\newblock \bibinfo{journal}{\emph{IEEE transactions on consumer electronics}}
  \bibinfo{volume}{38}, \bibinfo{number}{1} (\bibinfo{year}{1992}).
\newblock


\bibitem[Wang et~al\mbox{.}(2020a)]%
        {WangGBHK020}
\bibfield{author}{\bibinfo{person}{Yaxing Wang}, \bibinfo{person}{Abel
  Gonzalez{-}Garcia}, \bibinfo{person}{David Berga}, \bibinfo{person}{Luis
  Herranz}, \bibinfo{person}{Fahad~Shahbaz Khan}, {and} \bibinfo{person}{Joost
  van~de Weijer}.} \bibinfo{year}{2020}\natexlab{a}.
\newblock \showarticletitle{MineGAN: Effective Knowledge Transfer From GANs to
  Target Domains With Few Images}. In \bibinfo{booktitle}{\emph{CVPR}}.
\newblock


\bibitem[Wang et~al\mbox{.}(2020b)]%
        {wang2020semi}
\bibfield{author}{\bibinfo{person}{Yaxing Wang}, \bibinfo{person}{Salman Khan},
  \bibinfo{person}{Abel Gonzalez-Garcia}, \bibinfo{person}{Joost van~de
  Weijer}, {and} \bibinfo{person}{Fahad~Shahbaz Khan}.}
  \bibinfo{year}{2020}\natexlab{b}.
\newblock \showarticletitle{Semi-supervised learning for few-shot
  image-to-image translation}. In \bibinfo{booktitle}{\emph{CVPR}}.
\newblock


\bibitem[Weissenborn et~al\mbox{.}(2019)]%
        {weissenborn2019scaling}
\bibfield{author}{\bibinfo{person}{Dirk Weissenborn}, \bibinfo{person}{Oscar
  T{\"a}ckstr{\"o}m}, {and} \bibinfo{person}{Jakob Uszkoreit}.}
  \bibinfo{year}{2019}\natexlab{}.
\newblock \showarticletitle{Scaling Autoregressive Video Models}. In
  \bibinfo{booktitle}{\emph{ICLR}}.
\newblock


\bibitem[Xingjian et~al\mbox{.}(2015)]%
        {xingjian2015convolutional}
\bibfield{author}{\bibinfo{person}{SHI Xingjian}, \bibinfo{person}{Zhourong
  Chen}, \bibinfo{person}{Hao Wang}, \bibinfo{person}{Dit-Yan Yeung},
  \bibinfo{person}{Wai-Kin Wong}, {and} \bibinfo{person}{Wang-chun Woo}.}
  \bibinfo{year}{2015}\natexlab{}.
\newblock \showarticletitle{Convolutional LSTM network: A machine learning
  approach for precipitation nowcasting}. In
  \bibinfo{booktitle}{\emph{NeurIPS}}.
\newblock


\bibitem[Yang et~al\mbox{.}(2020b)]%
        {yang2020learning}
\bibfield{author}{\bibinfo{person}{Fuzhi Yang}, \bibinfo{person}{Huan Yang},
  \bibinfo{person}{Jianlong Fu}, \bibinfo{person}{Hongtao Lu}, {and}
  \bibinfo{person}{Baining Guo}.} \bibinfo{year}{2020}\natexlab{b}.
\newblock \showarticletitle{Learning texture transformer network for image
  super-resolution}. In \bibinfo{booktitle}{\emph{CVPR}}.
\newblock


\bibitem[Yang et~al\mbox{.}(2020a)]%
        {yang2020dpgn}
\bibfield{author}{\bibinfo{person}{Ling Yang}, \bibinfo{person}{Liangliang Li},
  \bibinfo{person}{Zilun Zhang}, \bibinfo{person}{Xinyu Zhou},
  \bibinfo{person}{Erjin Zhou}, {and} \bibinfo{person}{Yu Liu}.}
  \bibinfo{year}{2020}\natexlab{a}.
\newblock \showarticletitle{DPGN: Distribution Propagation Graph Network for
  Few-shot Learning}. In \bibinfo{booktitle}{\emph{CVPR}}.
\newblock


\bibitem[Zhang et~al\mbox{.}(2020)]%
        {zhang2020deepemd}
\bibfield{author}{\bibinfo{person}{Chi Zhang}, \bibinfo{person}{Yujun Cai},
  \bibinfo{person}{Guosheng Lin}, {and} \bibinfo{person}{Chunhua Shen}.}
  \bibinfo{year}{2020}\natexlab{}.
\newblock \showarticletitle{DeepEMD: Few-Shot Image Classification With
  Differentiable Earth Mover's Distance and Structured Classifiers}. In
  \bibinfo{booktitle}{\emph{CVPR}}.
\newblock


\bibitem[Zhang et~al\mbox{.}(2018)]%
        {zhang2018unreasonable}
\bibfield{author}{\bibinfo{person}{Richard Zhang}, \bibinfo{person}{Phillip
  Isola}, \bibinfo{person}{Alexei~A Efros}, \bibinfo{person}{Eli Shechtman},
  {and} \bibinfo{person}{Oliver Wang}.} \bibinfo{year}{2018}\natexlab{}.
\newblock \showarticletitle{The unreasonable effectiveness of deep features as
  a perceptual metric}. In \bibinfo{booktitle}{\emph{CVPR}}.
\newblock


\end{thebibliography}



\begin{thebibliography}{21}


\ifx \showCODEN    \undefined \def \showCODEN     #1{\unskip}     \fi
\ifx \showDOI      \undefined \def \showDOI       #1{#1}\fi
\ifx \showISBNx    \undefined \def \showISBNx     #1{\unskip}     \fi
\ifx \showISBNxiii \undefined \def \showISBNxiii  #1{\unskip}     \fi
\ifx \showISSN     \undefined \def \showISSN      #1{\unskip}     \fi
\ifx \showLCCN     \undefined \def \showLCCN      #1{\unskip}     \fi
\ifx \shownote     \undefined \def \shownote      #1{#1}          \fi
\ifx \showarticletitle \undefined \def \showarticletitle #1{#1}   \fi
\ifx \showURL      \undefined \def \showURL       {\relax}        \fi
\providecommand\bibfield[2]{#2}
\providecommand\bibinfo[2]{#2}
\providecommand\natexlab[1]{#1}
\providecommand\showeprint[2][]{arXiv:#2}

\bibitem[Chen et~al\mbox{.}(2018)]%
        {chen2018pixelsnail}
\bibfield{author}{\bibinfo{person}{Xi Chen}, \bibinfo{person}{Nikhil Mishra},
  \bibinfo{person}{Mostafa Rohaninejad}, {and} \bibinfo{person}{Pieter
  Abbeel}.} \bibinfo{year}{2018}\natexlab{}.
\newblock \showarticletitle{Pixelsnail: An improved autoregressive generative
  model}. In \bibinfo{booktitle}{\emph{ICML}}.
\newblock


\bibitem[Deng et~al\mbox{.}(2009)]%
        {deng2009imagenet}
\bibfield{author}{\bibinfo{person}{Jia Deng}, \bibinfo{person}{Wei Dong},
  \bibinfo{person}{Richard Socher}, \bibinfo{person}{Li-Jia Li},
  \bibinfo{person}{Kai Li}, {and} \bibinfo{person}{Li Fei-Fei}.}
  \bibinfo{year}{2009}\natexlab{}.
\newblock \showarticletitle{Imagenet: A large-scale hierarchical image
  database}. In \bibinfo{booktitle}{\emph{CVPR}}.
\newblock


\bibitem[Esser et~al\mbox{.}(2021)]%
        {esser2021taming}
\bibfield{author}{\bibinfo{person}{Patrick Esser}, \bibinfo{person}{Robin
  Rombach}, {and} \bibinfo{person}{Bjorn Ommer}.}
  \bibinfo{year}{2021}\natexlab{}.
\newblock \showarticletitle{Taming transformers for high-resolution image
  synthesis}. In \bibinfo{booktitle}{\emph{CVPR}}.
\newblock


\bibitem[Hong et~al\mbox{.}(2022)]%
        {hong2020deltagan}
\bibfield{author}{\bibinfo{person}{Yan Hong}, \bibinfo{person}{Li Niu},
  \bibinfo{person}{Jianfu Zhang}, {and} \bibinfo{person}{Liqing Zhang}.}
  \bibinfo{year}{2022}\natexlab{}.
\newblock \showarticletitle{Delta{GAN}: Towards diverse few-shot image
  generation with sample-specific delta}.
\newblock \bibinfo{journal}{\emph{ECCV}} (\bibinfo{year}{2022}).
\newblock


\bibitem[Hong et~al\mbox{.}(2020)]%
        {hong2020f2gan}
\bibfield{author}{\bibinfo{person}{Yan Hong}, \bibinfo{person}{Li Niu},
  \bibinfo{person}{Jianfu Zhang}, \bibinfo{person}{Weijie Zhao},
  \bibinfo{person}{Chen Fu}, {and} \bibinfo{person}{Liqing Zhang}.}
  \bibinfo{year}{2020}\natexlab{}.
\newblock \showarticletitle{F2GAN: Fusing-and-Filling GAN for Few-shot Image
  Generation}. In \bibinfo{booktitle}{\emph{ACM MM}}.
\newblock


\bibitem[Kaiser et~al\mbox{.}(2018)]%
        {kaiser2018fast}
\bibfield{author}{\bibinfo{person}{Lukasz Kaiser}, \bibinfo{person}{Samy
  Bengio}, \bibinfo{person}{Aurko Roy}, \bibinfo{person}{Ashish Vaswani},
  \bibinfo{person}{Niki Parmar}, \bibinfo{person}{Jakob Uszkoreit}, {and}
  \bibinfo{person}{Noam Shazeer}.} \bibinfo{year}{2018}\natexlab{}.
\newblock \showarticletitle{Fast decoding in sequence models using discrete
  latent variables}. In \bibinfo{booktitle}{\emph{ICML}}.
\newblock


\bibitem[Li et~al\mbox{.}(2020)]%
        {li2020few}
\bibfield{author}{\bibinfo{person}{Yijun Li}, \bibinfo{person}{Richard Zhang},
  \bibinfo{person}{Jingwan Lu}, {and} \bibinfo{person}{Eli Shechtman}.}
  \bibinfo{year}{2020}\natexlab{}.
\newblock \showarticletitle{Few-shot image generation with elastic weight
  consolidation}. In \bibinfo{booktitle}{\emph{NeurIPS}}.
\newblock


\bibitem[Liu et~al\mbox{.}(2019)]%
        {liu2019few}
\bibfield{author}{\bibinfo{person}{Ming{-}Yu Liu}, \bibinfo{person}{Xun Huang},
  \bibinfo{person}{Arun Mallya}, \bibinfo{person}{Tero Karras},
  \bibinfo{person}{Timo Aila}, \bibinfo{person}{Jaakko Lehtinen}, {and}
  \bibinfo{person}{Jan Kautz}.} \bibinfo{year}{2019}\natexlab{}.
\newblock \showarticletitle{Few-Shot Unsupervised Image-to-Image Translation}.
  In \bibinfo{booktitle}{\emph{ICCV}}.
\newblock


\bibitem[Nilsback and Zisserman(2008)]%
        {nilsback2008automated}
\bibfield{author}{\bibinfo{person}{Maria-Elena Nilsback} {and}
  \bibinfo{person}{Andrew Zisserman}.} \bibinfo{year}{2008}\natexlab{}.
\newblock \showarticletitle{Automated flower classification over a large number
  of classes}. In \bibinfo{booktitle}{\emph{CVGIP}}.
\newblock


\bibitem[Ojha et~al\mbox{.}(2021)]%
        {ojha2021few}
\bibfield{author}{\bibinfo{person}{Utkarsh Ojha}, \bibinfo{person}{Yijun Li},
  \bibinfo{person}{Jingwan Lu}, \bibinfo{person}{Alexei~A Efros},
  \bibinfo{person}{Yong~Jae Lee}, \bibinfo{person}{Eli Shechtman}, {and}
  \bibinfo{person}{Richard Zhang}.} \bibinfo{year}{2021}\natexlab{}.
\newblock \showarticletitle{Few-shot Image Generation via Cross-domain
  Correspondence}. In \bibinfo{booktitle}{\emph{CVPR}}.
\newblock


\bibitem[Peng et~al\mbox{.}(2021)]%
        {peng2021generating}
\bibfield{author}{\bibinfo{person}{Jialun Peng}, \bibinfo{person}{Dong Liu},
  \bibinfo{person}{Songcen Xu}, {and} \bibinfo{person}{Houqiang Li}.}
  \bibinfo{year}{2021}\natexlab{}.
\newblock \showarticletitle{Generating Diverse Structure for Image Inpainting
  With Hierarchical VQ-VAE}. In \bibinfo{booktitle}{\emph{CVPR}}.
\newblock


\bibitem[Radford et~al\mbox{.}(2019)]%
        {radford2019language}
\bibfield{author}{\bibinfo{person}{Alec Radford}, \bibinfo{person}{Jeffrey Wu},
  \bibinfo{person}{Rewon Child}, \bibinfo{person}{David Luan},
  \bibinfo{person}{Dario Amodei}, \bibinfo{person}{Ilya Sutskever},
  {et~al\mbox{.}}} \bibinfo{year}{2019}\natexlab{}.
\newblock \showarticletitle{Language models are unsupervised multitask
  learners}.
\newblock  (\bibinfo{year}{2019}).
\newblock


\bibitem[Ramachandran et~al\mbox{.}(2019)]%
        {ramachandran2019fast}
\bibfield{author}{\bibinfo{person}{Prajit Ramachandran}, \bibinfo{person}{Tom
  Le~Paine}, \bibinfo{person}{Pooya Khorrami}, \bibinfo{person}{Mohammad
  Babaeizadeh}, \bibinfo{person}{Shiyu Chang}, \bibinfo{person}{Yang Zhang},
  \bibinfo{person}{Mark Hasegawa-Johnson}, \bibinfo{person}{Roy Campbell},
  {and} \bibinfo{person}{Thomas Huang}.} \bibinfo{year}{2019}\natexlab{}.
\newblock \showarticletitle{Fast generation for convolutional autoregressive
  models}. In \bibinfo{booktitle}{\emph{ICLR}}.
\newblock


\bibitem[Razavi et~al\mbox{.}(2019)]%
        {razavi2019generating}
\bibfield{author}{\bibinfo{person}{Ali Razavi}, \bibinfo{person}{Aaron van~den
  Oord}, {and} \bibinfo{person}{Oriol Vinyals}.}
  \bibinfo{year}{2019}\natexlab{}.
\newblock \showarticletitle{Generating diverse high-fidelity images with
  vq-vae-2}. In \bibinfo{booktitle}{\emph{NeurIPS}}.
\newblock


\bibitem[Robb et~al\mbox{.}(2020)]%
        {robb2020few}
\bibfield{author}{\bibinfo{person}{Esther Robb}, \bibinfo{person}{Wen-Sheng
  Chu}, \bibinfo{person}{Abhishek Kumar}, {and} \bibinfo{person}{Jia-Bin
  Huang}.} \bibinfo{year}{2020}\natexlab{}.
\newblock \showarticletitle{Few-shot adaptation of generative adversarial
  networks}.
\newblock \bibinfo{journal}{\emph{arXiv preprint arXiv:2010.11943}}
  (\bibinfo{year}{2020}).
\newblock


\bibitem[Saito et~al\mbox{.}(2020)]%
        {saito2020coco}
\bibfield{author}{\bibinfo{person}{Kuniaki Saito}, \bibinfo{person}{Kate
  Saenko}, {and} \bibinfo{person}{Ming-Yu Liu}.}
  \bibinfo{year}{2020}\natexlab{}.
\newblock \showarticletitle{Coco-funit: Few-shot unsupervised image translation
  with a content conditioned style encoder}. In
  \bibinfo{booktitle}{\emph{ECCV}}.
\newblock


\bibitem[Van~Horn et~al\mbox{.}(2015)]%
        {van2015building}
\bibfield{author}{\bibinfo{person}{Grant Van~Horn}, \bibinfo{person}{Steve
  Branson}, \bibinfo{person}{Ryan Farrell}, \bibinfo{person}{Scott Haber},
  \bibinfo{person}{Jessie Barry}, \bibinfo{person}{Panos Ipeirotis},
  \bibinfo{person}{Pietro Perona}, {and} \bibinfo{person}{Serge Belongie}.}
  \bibinfo{year}{2015}\natexlab{}.
\newblock \showarticletitle{Building a bird recognition app and large scale
  dataset with citizen scientists: The fine print in fine-grained dataset
  collection}. In \bibinfo{booktitle}{\emph{CVPR}}.
\newblock


\bibitem[Vinyals et~al\mbox{.}(2016)]%
        {vinyals2016matching}
\bibfield{author}{\bibinfo{person}{Oriol Vinyals}, \bibinfo{person}{Charles
  Blundell}, \bibinfo{person}{Timothy Lillicrap}, \bibinfo{person}{Wierstra},
  {and} \bibinfo{person}{Daan}.} \bibinfo{year}{2016}\natexlab{}.
\newblock \showarticletitle{Matching networks for one shot learning}. In
  \bibinfo{booktitle}{\emph{NeurIPS}}.
\newblock


\bibitem[Wang et~al\mbox{.}(2020a)]%
        {WangGBHK020}
\bibfield{author}{\bibinfo{person}{Yaxing Wang}, \bibinfo{person}{Abel
  Gonzalez{-}Garcia}, \bibinfo{person}{David Berga}, \bibinfo{person}{Luis
  Herranz}, \bibinfo{person}{Fahad~Shahbaz Khan}, {and} \bibinfo{person}{Joost
  van~de Weijer}.} \bibinfo{year}{2020}\natexlab{a}.
\newblock \showarticletitle{MineGAN: Effective Knowledge Transfer From GANs to
  Target Domains With Few Images}. In \bibinfo{booktitle}{\emph{CVPR}}.
\newblock


\bibitem[Wang et~al\mbox{.}(2020b)]%
        {wang2020semi}
\bibfield{author}{\bibinfo{person}{Yaxing Wang}, \bibinfo{person}{Salman Khan},
  \bibinfo{person}{Abel Gonzalez-Garcia}, \bibinfo{person}{Joost van~de
  Weijer}, {and} \bibinfo{person}{Fahad~Shahbaz Khan}.}
  \bibinfo{year}{2020}\natexlab{b}.
\newblock \showarticletitle{Semi-supervised learning for few-shot
  image-to-image translation}. In \bibinfo{booktitle}{\emph{CVPR}}.
\newblock


\bibitem[Yu et~al\mbox{.}(2021)]%
        {yu2021diverse}
\bibfield{author}{\bibinfo{person}{Yingchen Yu}, \bibinfo{person}{Fangneng
  Zhan}, \bibinfo{person}{Rongliang Wu}, \bibinfo{person}{Jianxiong Pan},
  \bibinfo{person}{Kaiwen Cui}, \bibinfo{person}{Shijian Lu},
  \bibinfo{person}{Feiying Ma}, \bibinfo{person}{Xuansong Xie}, {and}
  \bibinfo{person}{Chunyan Miao}.} \bibinfo{year}{2021}\natexlab{}.
\newblock \showarticletitle{Diverse image inpainting with bidirectional and
  autoregressive transformers}.
\newblock \bibinfo{journal}{\emph{arXiv preprint arXiv:2104.12335}}
  (\bibinfo{year}{2021}).
\newblock


\end{thebibliography}

\end{document}


\title{Supplementary of Few-shot Image Generation Using Discrete Content Representation}

\acmSubmissionID{1685}





\settopmatter{authorsperrow=4}
\author{Yan Hong}
\affiliation{%
  \institution{MoE Key Lab of Artificial Intelligence, Shanghai Jiao Tong University}
  \country{China}
}
\email{yanhong.sjtu@gmail.com}

\author{Li Niu}
\authornote{Corresponding author.}
\affiliation{%
  \institution{MoE Key Lab of Artificial Intelligence, Shanghai Jiao Tong University}
    \country{China}
}
\email{ustcnewly@sjtu.edu.cn}

\author{Jianfu Zhang}
\affiliation{%
  \institution{MoE Key Lab of Artificial Intelligence, Shanghai Jiao Tong University}
  \country{China}
}
\email{c.sis@sjtu.edu.cn}



\author{Liqing Zhang}
\affiliation{\institution{MoE Key Lab of Artificial Intelligence, Shanghai Jiao Tong University}
\country{China}
}
\email{zhang-lq@cs.sjtu.edu.cn}

%

\renewcommand{\shortauthors}{Yan Hong, Li Niu, Jianfu Zhang, \& Liqing Zhang}

\maketitle
In this document, we provide additional material to support our main submission. 
In Section~\ref{sec:implementary_details}, we describe the details of three datasets and network structure.
In Section~\ref{sec:dictionary}, we study the effects of dictionary of local content vectors $\mathcal{D}$ including the content map size $N$ and dictionary length $K$.
In Section~\ref{sec:transformer_blocks}, we report the results of our transformer-based content map generator with different numbers of blocks.
In Section~\ref{sec:more_generation_comparison}, we visualize more images generated by Disco-FUNIT (1st stage) (\emph{resp.,} Disco-FUNIT), and also compare with few-shot image translation (\emph{resp.,} generation) baselines. 
In Section~\ref{sec:few_shot_ability},  we compare our method with baselines in $H$-shot setting with different $H$.
In Section~\ref{sec:comparison_with_finetune_baseline}, we report the comparison results between Disco-FUNIT and other few-shot image generation method relying on finetuning process in the test phase. In Section~\ref{sec:cocofunit_1st_stage}, we replace FUNIT with COCO-FUNIT in our method. In Section~\ref{sec:discussion_vqgan}, we discuss the difference between our method and VQGAN.
In Section~\ref{sec:limitation}, we provide some failure cases and limitation analyses during testing.
Recall that our method has two training stages. After the first training stage, ``Disco-FUNIT (1st stage)" is a quantized few-shot image translation method, which still requires seen images to provide content maps. After the second training stage, ``Disco-FUNIT" becomes a few-shot image generation method, which can sample diverse content maps conditioned on the style vector. In the remainder of this document, we will use ``Disco-FUNIT (1st stage)" and ``Disco-FUNIT" to distinguish these two stages.

\section{Dataset Details and Network Architectures}\label{sec:implementary_details}
We use Pytorch $1.7.0$ to implement our model, which is distributed on RTX 2080 Ti GPU. 

\noindent\textbf{Datasets}
Our method is evaluated on three datasets:  Flowers~\cite{nilsback2008automated}, Animal Faces~\cite{deng2009imagenet}, and NABirds~\cite{van2015building}. We summarize the number of seen (\emph{resp.,} unseen) categories and seen (\emph{resp.,} unseen) images in Table~\ref{tab:dataset}.

\noindent\textbf{Disentanglement Framework with Quantization} 
The architecture of style encoder $E_s$, content encoder $E_c$, decoder $F$, and multi-class discriminator $D$ used in our experiments are the same as FUNIT~\cite{liu2019few}. The only modification in our paper is to add two convolutional layers in content encoder $E_c$ for quantizing continuous content maps into discrete content maps. In detail, based on the content encoder of FUNIT, we add two $1 \times 1 \times d_c$ convolutional layers before and after quantizing continuous content map $\bm{C}^{x} \in \mathcal{R}^{w\times h\times {d_c}}$ into discrete content map $\hat{\bm{C}}^{x} \in \mathcal{R}^{w\times h\times {d_c}}$, respectively. In our experiment, the resolution of content image $\bm{x}$ and style image $\bm{y}$ are $128\times 128$. We set $w\times h=16\times 16$ and $d_c=512$. The size of content maps $w \times h$ can be tuned by varying the number of downsampling layers in content encoder $E_c$.
When quantizing continuous content maps $\bm{C}^x$ into discrete content map $\hat{\bm{C}}^x$, we learn a dictionary of local content vectors $\mathcal{D} \in \mathcal{R} ^{K\times d_c}$, where $K$ is the length of dictionary and set as $1024$ in our experiment.
The storage size of a dictionary is $0.53$MB, which is roughly equivalent to that of 12 images with size $128\times 128$. In contrast, if we store a large number of seen images, \emph{e.g.}, 10000 images, the storage size is $390.61$MB and the diversity is limited by the number of seen images. Compared with few-shot image generation such as FUNIT and COCO-FUNIT, quantized few-shot image translation method “Disco-FUNIT (1st stage)”  only saves the indices of content maps of seen images based on the learned dictionary instead of storing seen images.
Moreover, quantization can help remove redundant and noisy information from content maps, contributing to better disentanglement (see Section~\ref{sec:more_generation_comparison}). 


\begin{table}[t]
  \caption{The splits of seen/unseen images (``img'') and categories (``cat'') on three datasets.} 
  \centering
  \resizebox{0.7\columnwidth}{!} {
  \begin{tabular}{l|rr|rr}  
    \toprule[0.8pt]
     \multirow{2}{*}{Dataset}& \multicolumn{2}{c|}{Seen} &\multicolumn{2}{c}{Unseen}\cr
      &\#img & \#cat   &\#img & \#cat \cr
     \hline
     Flowers & 7121&85 & 1068&17 \cr
     Animal Faces & 96621&119 & 20863&30   \cr
     NABirds & 38306&444 &  10221&111 \cr
    \bottomrule[0.8pt]
  \end{tabular}
  }
  \label{tab:dataset}
\end{table}
\begin{figure}[t]
\begin{center}
\includegraphics[scale=0.35]{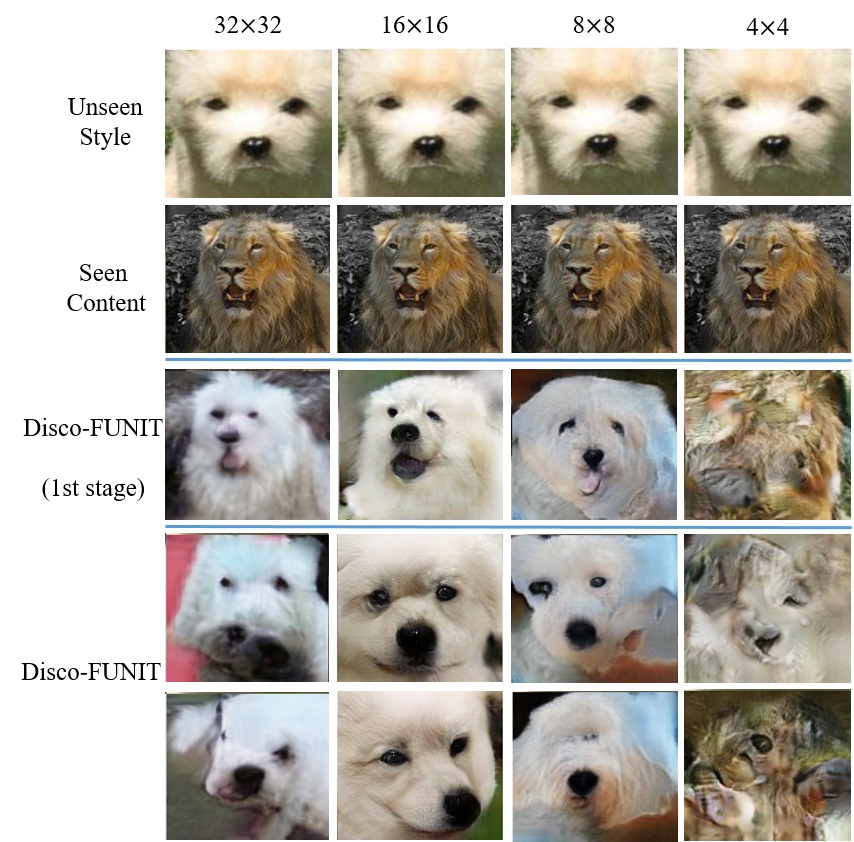}
\end{center}
\caption{Images generated by Disco-FUNIT (1st stage) and Disco-FUNIT with different sizes of content maps $w \times h$. From top to bottom: unseen style image, seen content image, image generated by Disco-FUNIT (1st stage), and two images generated from Disco-FUNIT by sampling two content maps (without using content images in row 2).}
\label{fig:content_map_size} 
\end{figure}



\noindent\textbf{Content Map Generator} 
The architecture of our content map generator $T_{c|s}$ is identical to the GPT2 architecture~\cite{radford2019language} with input size of $1 + 16 \times 16 $ for the conditional style vector $\bm{s}^y$ and the sequence of discrete content map $\{\hat{\bm{c}}^x_i|_{i=1}^{256}\}$. The only modification is to add one $d_s \times K$ fully-connected layer to encode conditional style vector $\bm{s}^{x} \in \mathcal{R}^{{d_s}}$ to the same size of one-hot token embedding $\mathcal{R}^{{K}}$, where $d_s=64$  and $K=1024$ is the length of dictionary of local content vectors. The capacity of content map generator can be tuned by varying the number $l$ of transformer blocks, which is set as $l=12$ in our experiment. During inference, we use $T_{c|s}$ in a sliding-window manner with a window size of $3$ to speed up sampling, which is similar to~\cite{esser2021taming}. We adopt top-$k$ sampling strategy to randomly sample from the $k$ ($k$ is 100 in our experiments) most likely next local content vectors. 

\begin{table}[t]
  \caption{Performance evaluation of the images generated by Disco-FUNIT(1st stage)/Disco-FUNIT when using different sizes of content maps $w \times h$ on Animal Faces dataset.} 
  \centering
  \resizebox{1\columnwidth}{!} {
  \begin{tabular}{l|r|r|r}  
    \hline
     $w \times h$&  FID $\downarrow$  & LPIPS $\uparrow$ & Accuracy(\%) $\uparrow$ \cr
    \hline
    $4\times4$& 	340.35/335.47 &0.3991/0.4104 & -/-\cr
    $8\times8$ &97.81/92.80 & 0.4471/0.4182&56.45/56.08\cr
    $16\times16$ & $\textbf{75.47/71.44}$ &$\textbf{0.4423/0.4511}$  & $\textbf{60.96/61.85}$ \cr
    $32\times32$ & 93.12/111.07 &0.4718/0.4631 & 57.89/55.42\cr
    \hline
  \end{tabular}
  }
  \label{tab:content_map}
\end{table}


\section{Investigation of Dictionary of Local Content Vectors} \label{sec:dictionary}
In this section, we evaluate the effects of content map size $w\times h$ and the length of dictionary of local content vectors $K$. By taking  Animal Faces dataset as an example, we report FID and LPIPS in one-shot setting as well as 10way-1shot few-shot classification accuracy as in Section 5.2 in the main paper. 

\noindent\textbf{Effect of Content Map Size} To investigate the effect of content maps size $N=w\times h$, we vary the number of downsampling blocks in the content encoder. Content encoder $E_c$ encodes input content images of size $128\times 128$ into discrete content maps of size $w\times h$. We obtain $32\times 32$ (\emph{resp.,} $16\times 16$, $8\times 8$, and $4\times 4$) discrete content maps by setting the number of downsampling layers as $2$ (\emph{resp.,} $3$, $4$, and $5$). We visualize the unseen image providing style vector, the seen image providing content map, the generated images from Disco-FUNIT (1st stage) and Disco-FUNIT in Figure~\ref{fig:content_map_size}. We also report the quantitative results in Table~\ref{tab:content_map}. For discrete content map with large size $32\times 32$, the appearance of generated image from Disco-FUNIT (1st stage) looks quite different from that of style image, which can be explained as below. Discrete content maps with large size have small receptive field and fail to remove the global noisy information when quantizing continuous content map into discrete content map, leading to insufficient disentanglement. Although the content information of generated image belongs to seen image (\emph{i.e.}, the pose of animal face),  the discrete content map with large size $32\times 32$ also carries other noisy information which could degrade the performance of disentanglement. For Disco-FUNIT, the content map generator also fails to sample content maps compatible with the style vector, resulting in low-quality images. Considering the poor quality of generated images for discrete content map with size $4\times 4$, we omit its low 10way-1shot accuracy in Table~\ref{tab:content_map}.
For discrete content map with small size $8\times 8$, the overall structure of generated image from Disco-FUNIT (1st stage) is similar to the content image, but some local structures are distorted. The discrete content map with size $4\times 4$ even fails to generate plausible images. 

In contrast, content map size $16\times 16$ can translate seen image to unseen image well, which indicates that content map with size $16\times 16$ contains full content information of seen image and removes redundant noisy information. On the basis of disentanglement with content map quantization, content map generator $T_{c|s}$ can also sample discrete content maps compatible with the style vector to produce diverse and realistic images.
Hence, learning perceptually rich content maps can help our disentanglement model learn disentangled content maps and style vectors sufficiently, and produce perceptually faithful  images without distortion. Only with perceptually rich content maps, the dictionary of local content vectors $\mathcal{D}$ can be well-constructed, which determines the performance of content map generator $T_{c|s}$ for producing new images in the second stage.
\setlength{\tabcolsep}{5pt}
\begin{table}[t]
  \caption{Performance evaluation of the images generated by Disco-FUNIT(1st stage)/Disco-FUNIT when using different  dictionary lengths $K$ on Animal Faces dataset.} 
  \centering
  \resizebox{1\columnwidth}{!} {
  \begin{tabular}{l|r|r|r}  
    \hline
     K &  FID $\downarrow$  & LPIPS $\uparrow$ & Accuracy(\%) $\uparrow$ \cr
    \hline
    32 &102.88/108.56 &0.4966/0.4811& 52.84/52.26 \cr
    64 &100.53/107.49 &0.5032/0.4867 & 53.95/52.13\cr
    128 &82.16/84.92 & 0.4764/0.4404 & 57.49/55.26\cr
    256 & 95.82/92.23& 0.4939/0.4433 & 56.34/55.63 \cr
    512  &79.91/78.76 &0.4591/0.4091& 59.52/57.17    \cr
    1024& $\textbf{75.47/71.44}$ &$\textbf{0.4423/0.4511}$ & $\textbf{60.96/61.85}$  \cr
    2048 &80.87/83.92 & 0.4536/0.4472& 59.68/59.12 \cr
    8192 &86.64/91.45  &0.4801/0.4426 & 57.78/58.87  \cr
    \hline
  \end{tabular}
  }
  \label{tab:dictionary_length}
\end{table}

\begin{figure*}
\begin{center}
\includegraphics[scale=0.28]{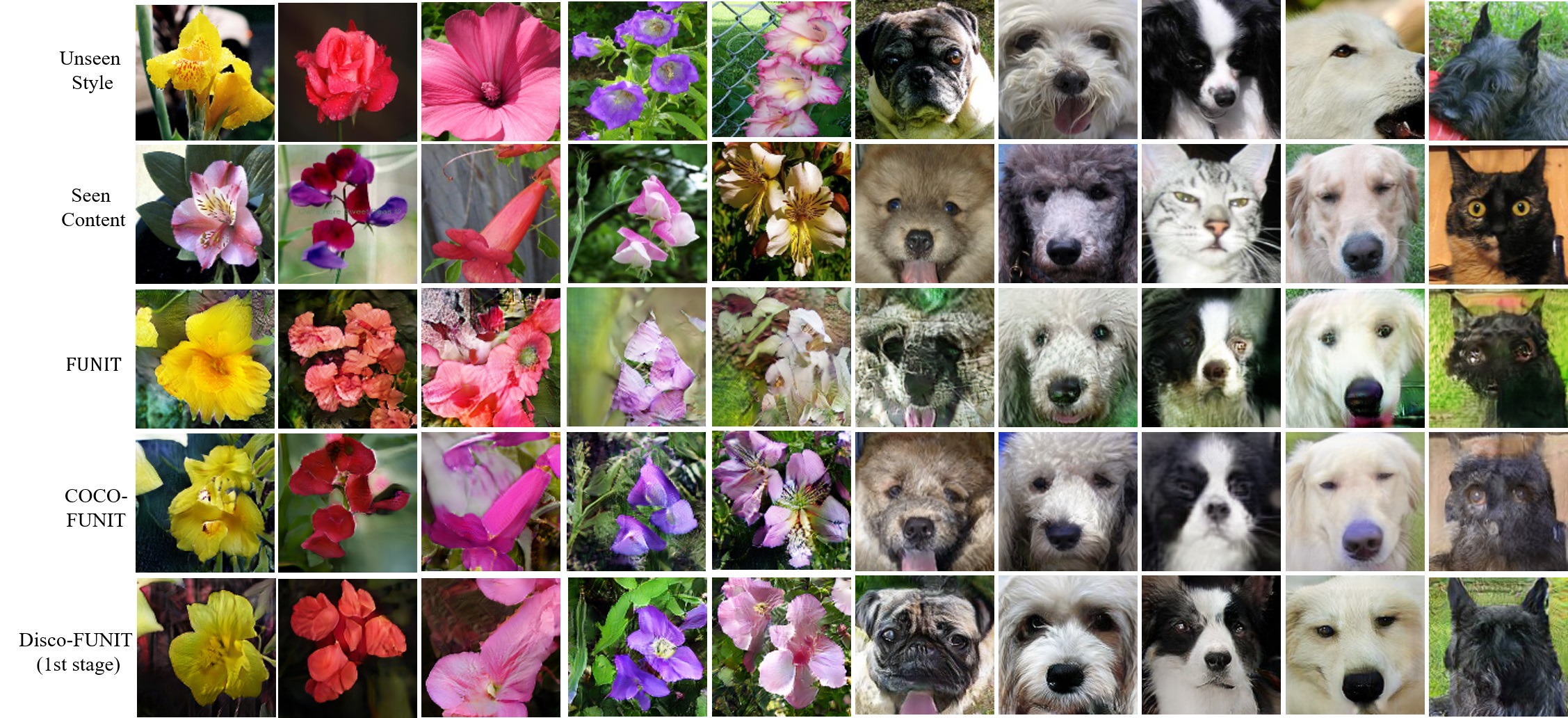}
\end{center}
\caption{Images generated by FUNIT, COCO-FUNIT, and Disco-FUNIT (1st stage) in 1-shot setting on Flowers and Animal Faces datasets. From top to bottom: unseen style image, seen content image, images generated by FUNIT, COCO-FUNIT, and Disco-FUNIT (1st stage).}
\label{fig:more_images_1stage_comparison} 
\end{figure*}

\noindent\textbf{Effect of Dictionary Length of Local Content Vectors}
To study the effect of dictionary of local content vectors, we tune the length of dictionary $K$ in $\{32, 64, 128, 512, 1024, 2048\}$ and report the evaluation results in Table~\ref{tab:dictionary_length}. From Table~\ref{tab:dictionary_length}, it can be seen that the small dictionary (\emph{i.e.}, $K=32, 64, 128, 256$) results in high FID values and high LPIPS values. High FID values indicate the  distribution mismatch between generated images and real images. For high LPIPS values, we conjecture that local content vectors sampled from small dictionary may form distorted images, leading to large variance among generated images. Due to the limited capacity of small dictionary, the continuous content map may be divergent from its quantized discrete content maps, which leads to the loss of important content information and poor results from ``Disco-FUNIT (1st stage)"/``Disco-FUNIT".  Another observation is that large dictionary of local content vectors (\emph{i.e.}, $K=2048, 8192$) does not improve the quality of generated images from ``Disco-FUNIT (1st stage)"/``Disco-FUNIT", probably because large dictionary of local content vectors may maintain some redundant and noisy information when quantizing continuous content maps.

\begin{table}[t]
  \caption{Performance evaluation of the images generated by Disco-FUNIT when using different numbers of transformer blocks $l$ on Animal Faces dataset.} 
  \centering
  \resizebox{0.65\columnwidth}{!} {
  \begin{tabular}{l|r|r|r}  
    \hline
     $l$ &  FID $\downarrow$  & LPIPS $\uparrow$ &Accuracy(\%) $\uparrow$ \cr
    \hline
    $6$& 81.25	 &0.3789 & 54.18 \cr
    $12$ &$\textbf{71.44}$ &$\textbf{0.4511}$ & $\textbf{61.85}$\cr
    $24$ &74.56  & 0.4169 & 57.52 \cr
    \hline
  \end{tabular}
  }
  \label{tab:transformer_layers}
\end{table}

\begin{figure*}
\begin{center}
\includegraphics[scale=0.8]{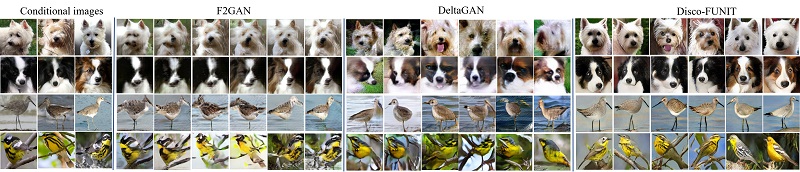}
\end{center}
\caption{Images generated by F2GAN~\cite{hong2020f2gan}, DeltaGAN~\cite{hong2020deltagan}, and our method Disco-FUNIT in 3-shot setting on two datasets ( Animal Faces in row 1-4 and NABirds in row 5-8). The conditional images are in the left three columns. The generated images are arranged based on the conditional image (two generated images for each conditional image).}
\label{fig:visualization_compare_supple} 
\end{figure*}

\begin{figure*}
\begin{center}
\subfigure[Flowers]{
\begin{minipage}[t]{1\linewidth}
\centering
\includegraphics[scale=0.24]{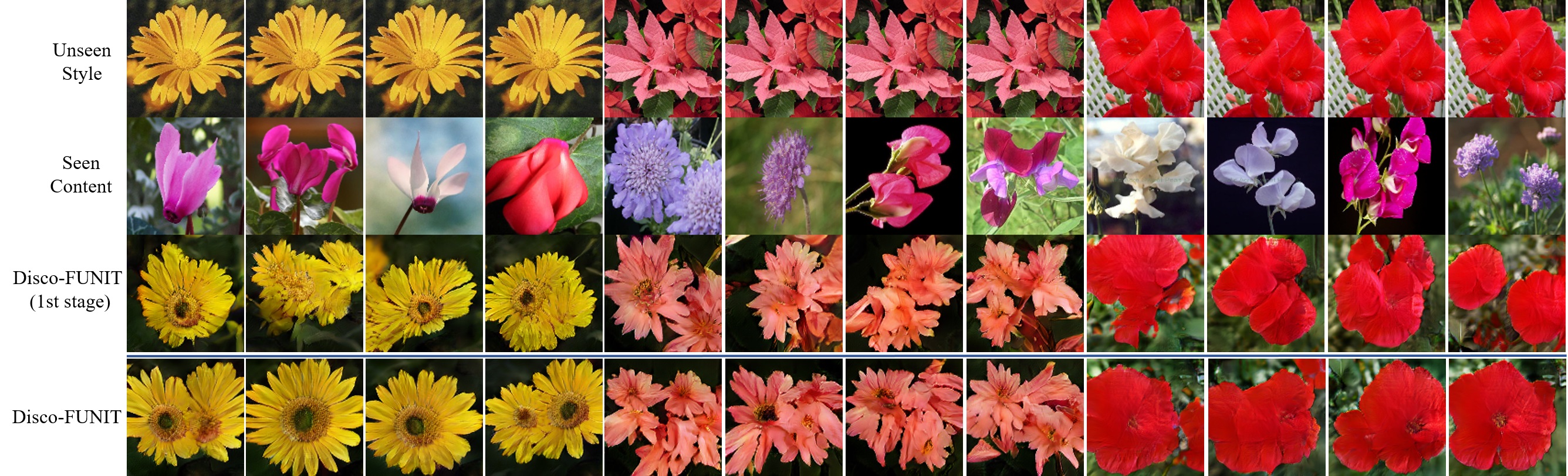}
\end{minipage}
}
\subfigure[Animal Faces]{
\begin{minipage}[t]{1\linewidth}
\centering
\includegraphics[scale=0.24]{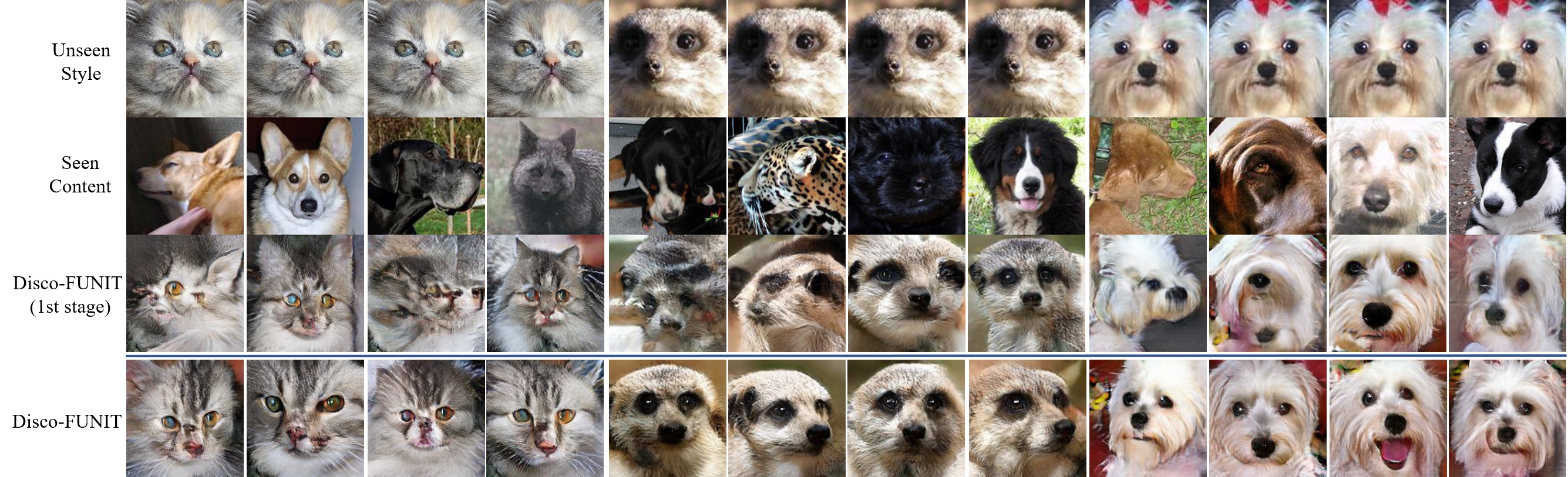}
\end{minipage}
}
\subfigure[NABirds]{
\begin{minipage}[t]{1\linewidth}
\centering
\includegraphics[scale=0.24]{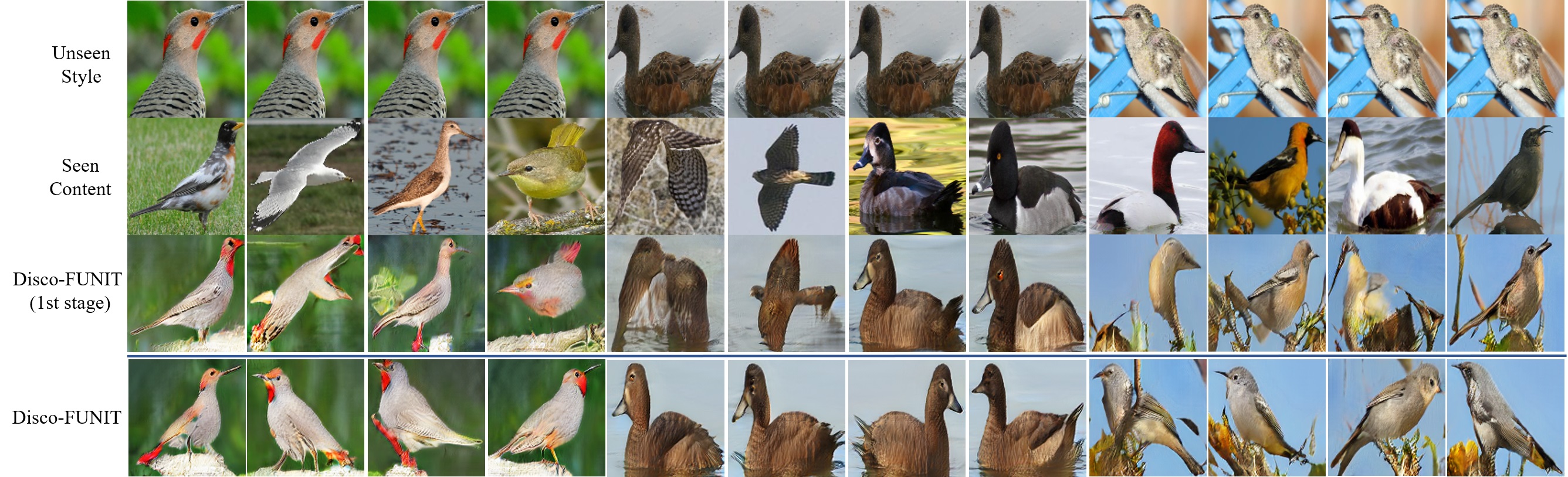}
\end{minipage}
}
\end{center}
\caption{Images generated by ``Disco-FUNIT (1st stage)" (\emph{resp.}, ``Disco-FUNIT") in 1-shot setting on Flowers, Animal Faces, and NABirds datasets. From top to bottom: unseen style image, seen content image, image generated by Disco-FUNIT (1st stage), and image generated from Disco-FUNIT by sampling content maps (without using content images in row 2). }
\label{fig:stage12_threedatasets} 
\end{figure*}

\begin{figure*}
\begin{center}
\includegraphics[scale=0.75]{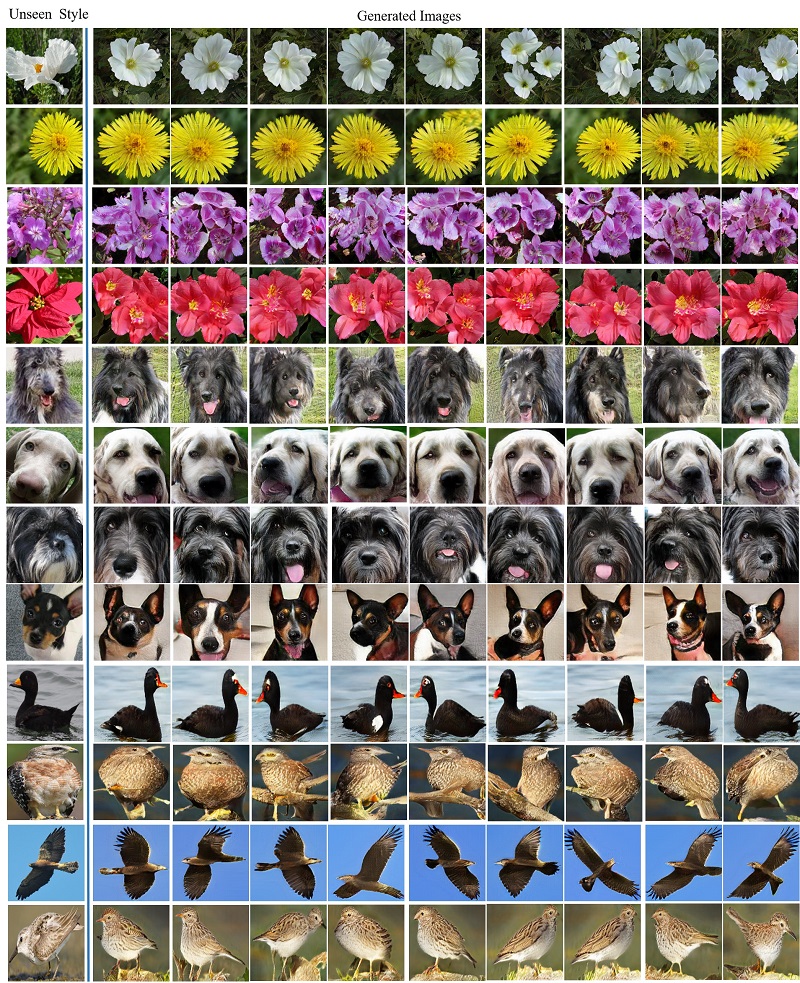}
\end{center}
\caption{Images generated by Disco-FUNIT in 1-shot setting on three datasets (from top to bottom:  Flowers in row 1-4, Animal Faces in row 5-8, and NABirds in row 9-12). The conditional images are in the leftmost column.}
\label{fig:more_images_2stage} 
\end{figure*}
\begin{figure}[ht]
\begin{center}
\includegraphics[scale=0.36]{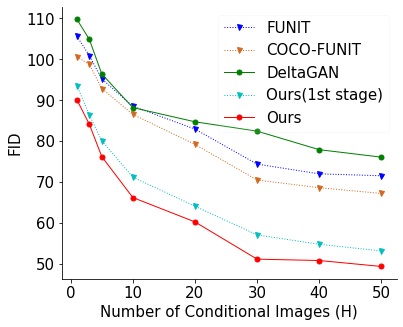}
\includegraphics[scale=0.36]{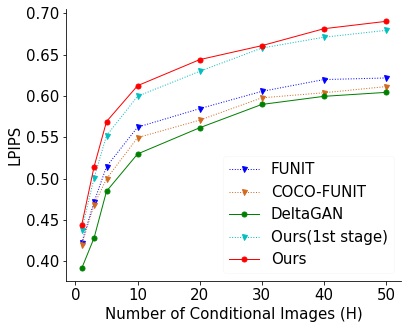}
\end{center}
\caption{FID and LPIPS comparison of DeltaGAN, Disco-FUNIT (1st stage), and our full method with different numbers (H) of conditional images on Flowers dataset.}
\label{fig:fid_curve} 
\end{figure}

In comparison, $K=1024$ is a suitable value, which can ensure the good performance of ``Disco-FUNIT (1st stage)"/``Disco-FUNIT". Specifically, the dictionary of local content vectors with $K=1024$ can maintain necessary content information and remove redundant noisy information, which can achieve better disentanglement and help extract purer content maps/style vectors. On the basis of well-constructed dictionary of local content vectors, our style-conditioned content map generator can sufficiently model the distribution of discrete content maps conditioned on style vector for producing diverse and realistic images.  

\section{Investigation of Transformer Block} \label{sec:transformer_blocks}
To evaluate the effect of the number of transformer blocks $l$, we vary $l$ in $\{6, 12,24\}$ on Animal Faces dataset. Similar to Section \ref{sec:dictionary}, we report FID, LPIPS, 
10way-1shot accuracy in Table~\ref{tab:transformer_layers}. We can see that $l=12$ achieves the lowest FID as well as the highest LPIPS and 10way-1shot accuracy, which indicates that transformer with $l=12$ can model the autoregressive distribution of discrete content maps for producing diverse and realistic images.

\section{More Visualization Results}\label{sec:more_generation_comparison}

\noindent\textbf{Comparison with Few-shot Image Translation Baselines}
We visualize some example images generated by FUNIT, COCO-FUNIT, and Disco-FUNIT (1st stage) on Flowers and Animal Faces datasets in Figure~\ref{fig:more_images_1stage_comparison}.
We randomly sample content images from seen categories and style images from unseen categories to produce images for few-shot image translation methods. From Figure~\ref{fig:more_images_1stage_comparison},  we can see that the images generated by FUNIT have a lot of artifacts (see column 3-6), whereas COCO-FUNIT can generate images with higher quality, but the generated images lose some details and look considerably different from the style image (see column 3, 6-7, and 9-10).  In contrast, the images generated by Disco-FUNIT (1st stage) are more realistic and maintain the details of unseen style images. Besides, images generated by Disco-FUNIT (1st stage) can also retain the pose of seen content images (see column $1$) without distortion. The comparison results demonstrate the benefit of learning a dictionary of local content vectors $\mathcal{D}$ via quantizing continuous content maps into discrete content maps. Discrete content maps formed from dictionary $\mathcal{D}$ can remove redundant and noisy information from continuous content maps to achieve better disentanglement.

\noindent\textbf{Comparison with Few-shot Image Generation Baselines}
We visualize more images generated by F2GAN, DeltaGAN, and our method in $3$-shot setting in Figure~\ref{fig:visualization_compare_supple}.  From comparison, we can see that the diversity of images generated by F2GAN is limited, due to insufficient fusion of conditional images. For DeltaGAN, generated images are relatively diverse, but their quality is inferior to Disco-FUNIT. In contrast, our method can produce diverse images with higher quality, which indicates that our discrete content maps can cooperate with disentanglement framework and the style-conditioned content map generator can sample more compatible content maps with style vector. 

\noindent\textbf{Comparison between Disco-FUNIT (1st stage) and Disco-FUNIT}
We compare the generated images from Disco-FUNIT (1st stage) and the generated images from Disco-FUNIT on Flowers, Animal Faces, and NABirds dataset in Figure~\ref{fig:stage12_threedatasets}. We can see that images from Disco-FUNIT (1st stage) can maintain the appearance of style image and adjust its pose to content image, which validate the effectiveness of our quantized few-shot image translation method. However, by comparing Disco-FUNIT (1st stage) and Disco-FUNIT, we can see that the quality of some images from Disco-FUNIT (1st stage) is worse than Disco-FUNIT due to unreasonable poses (see column 3, 5, 10 in subfigure (b) in Figure~\ref{fig:stage12_threedatasets}, and column 2, 5,6,9,11 in subfigure (c) in Figure~\ref{fig:stage12_threedatasets}). The reason is that some randomly selected content maps from seen images may be incompatible with the style vector of unseen image, leading to implausible generated images. In contrast, the discrete content maps sampled from our content map generator are compatible with the style vector for producing satisfactory images, which demonstrates that our style-conditioned content map autoregression stage can effectively solve the issue of incompatible content maps and style vectors.

To further demonstrate the effectiveness of our style-conditioned content map generator $T_{c|s}$, we use content map generator to sample diverse discrete content maps conditioned on an unseen style vector for generating new images, as shown in Figure~\ref{fig:more_images_2stage}. It can be seen that the images generated by Disco-FUNIT are diverse and realistic, because our content map generator can sample diverse discrete content maps compatible with the unseen style vector for producing images with reasonable content information (\emph{i.e.}, pose of animals and birds, structure of flowers).

\section{Few-shot Generation Ability}\label{sec:few_shot_ability}
Here, we repeat the experiments in Section 5.2 in the main paper except tuning $H$ in a wide range. Recall that $H$ in $H$-shot setting means that $H$ real images are provided for each unseen category. We use our full method, Disco-FUNIT (1st), competitive few-shot image generation baseline (\emph{i.e.}, DeltaGAN), and few-shot image translation baselines (\emph{i.e.}, FUNIT and COCO-FUNIT) to generate 128 new images for each  unseen  category in $H$-shot setting. We also adopt the same quantitative evaluation metrics (FID and LPIPS) to measure image quality and diversity as in Section 5.2 in the main paper. We plot the FID curve and LPIPS curve of five methods with increasing $H$ in Figure~\ref{fig:fid_curve}. It can be seen that Disco-FUNIT (\emph{resp.,} Disco-FUNIT (1st stage))  outperforms DeltaGAN (\emph{resp.,} FUNIT and COCO-FUNIT) by a large margin with all values of $H$, especially when $H$ is very small. 
These results demonstrate that our full method can generate abundant diverse and realistic images even if only a few (\emph{e.g.}, 10) real images are provided.

\section{Comparison with Other Few-shot Image Generation} \label{sec:comparison_with_finetune_baseline}
\begin{figure}[t]
\begin{center}
\includegraphics[scale=0.14]{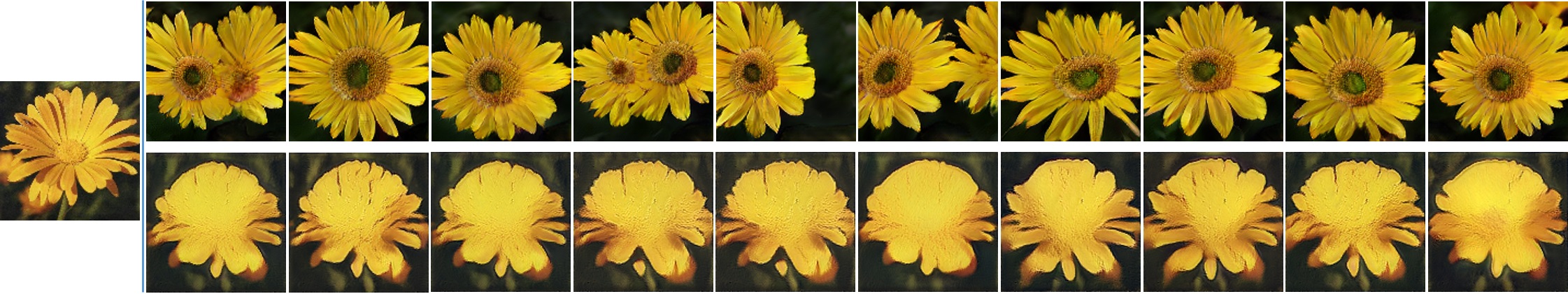}
\end{center}
\vspace{-0.5cm}
\caption{Images generated by \cite{ojha2021few} (bottom row) and our method (top row) conditioned on one image (leftmost column). }
\label{fig:visualization_compare_finetune} 
\end{figure}

As discussed in Section 2 in the main paper, our setting is quite different from recent works \cite{ojha2021few,li2020few,robb2020few,WangGBHK020}. Our method can achieve instant adaptation to multiple unseen categories without finetuning, while the abovementioned methods need to finetune the trained model for each unseen category, which is very resource-consuming and time-consuming. We compare the performance of \cite{ojha2021few} with our method on Flowers dataset.  For  \cite{ojha2021few}, we train a source model on all seen images during training. At test stage, we finetune the source model on each selected unseen category with one image (1-shot setting in Section 5.2 in main paper) and produce $128$ images by sampling random vectors for evaluation.
We report the results in Table{~\ref{tab:comparison_metric_finetune}} and visualize some example images generated by \cite{ojha2021few} and our method in Figure~{\ref{fig:visualization_compare_finetune}}. We can see that our method outperforms \cite{ojha2021few} and produces more diverse and realistic images.

\setlength{\tabcolsep}{4pt}
\begin{table}[t]
  \centering
  \resizebox{0.4\columnwidth}{!} {
  \begin{tabular}{l|c|c}  
    \hline
     Setting&  FID $\downarrow$  & LPIPS $\uparrow$ \cr
    \hline
     \cite{ojha2021few} &109.89 & 0.2879 \cr
    \hline
    Disco-FUNIT & \textbf{90.12} & \textbf{0.4436}   \cr
    \hline
  \end{tabular}
  }
\caption{FID ($\downarrow$) and LPIPS ($\uparrow$) of images generated by \cite{ojha2021few} and our method on Flowers dataset.} 
\label{tab:comparison_metric_finetune}
\end{table}
\setlength{\tabcolsep}{5pt}
\begin{table}[t]
  \centering
  \resizebox{0.9\columnwidth}{!} {
  \begin{tabular}{l|c|c|c}  
    \hline
     Setting&  FID $\downarrow$  & LPIPS $\uparrow$ & Accuracy(\%) $\uparrow$ \cr
    \hline
    FUNIT&105.65&0.4221&55.98 \cr
    COCO-FUNIT&100.61& 0.4197&57.18 \cr
    Disco-COCO (1st stage) & 97.28  &0.4286   &61.11  \cr
    Disco-FUNIT (1st stage) & \textit{93.59} & \textit{0.4373}  & \textit{62.59}  \cr
    \hline
    Disco-COCO &95.89  &0.4316   &61.94  \cr
    Disco-FUNIT & \textbf{90.12} & \textbf{ 0.4436}  & \textbf{63.11}  \cr
    \hline
  \end{tabular}
  }
\caption{Performance comparison between Disco-FUNIT and Disco-COCO in each stage in 1-shot setting on Flowers dataset.}
  \label{tab:compare_coco}
\end{table}

\section{Using Different Disentanglement Framework}\label{sec:cocofunit_1st_stage}
FUNIT~\cite{liu2019few} and COCO-FUNIT~\cite{saito2020coco} are two popular disentanglement frameworks in few-shot setting.
\textit{The reason of choosing FUNIT is the superior performance of quantizing the content maps extracted from FUNIT.} 
We replace FUNIT with COCO-FUNIT and the two-stage training process remains the same. 
We refer to full method (\emph{resp.,} first stage) in this configuration as Disco-COCO (\emph{resp.,} Disco-COCO (1st stage)).
The results in Table~{\ref{tab:compare_coco}} show that Disco-FUNIT outperforms Disco-COCO in each stage. The superior performance of Disco-COCO (1st stage) compared with COCO-FUNIT demonstrates the effectiveness of content map quantization. Another observation is that Disco-COCO outperforms Disco-COCO (1st stage), which demonstrates the effectiveness of our style-conditioned content map generator in second stage.

\section{Comparison with VQGAN}\label{sec:discussion_vqgan}
 Our proposed method is similar with VQGAN~\cite{esser2021taming} in terms of vector quantization and transformer-based autoregression. However, these techniques are not firstly proposed in VQGAN and have been widely used in different tasks as described in Section 2 in main paper. Similar to VQGAN, in ~\cite{kaiser2018fast}, transformer-based autoregression models are applied to achieve decoding using discrete latent variables learned with vector quantization technique.
\textit{Although VQGAN can be applied to class-conditioned image generation, it cannot produce images for unseen categories and thus inapplicable to our task}. 
We make the first attempt to quantize content map under disentanglement framework, which can save compact dictionary instead of amounts of seen images and prune redundant information to enhance disentanglement performance. Moreover, to solve the incompatibility issue between style vector and content map in FUNIT, we propose to model the autoregressive distribution of content map given a style vector, which has also not been explored before.

\begin{figure} [t]
\begin{center}
\includegraphics[scale=0.2]{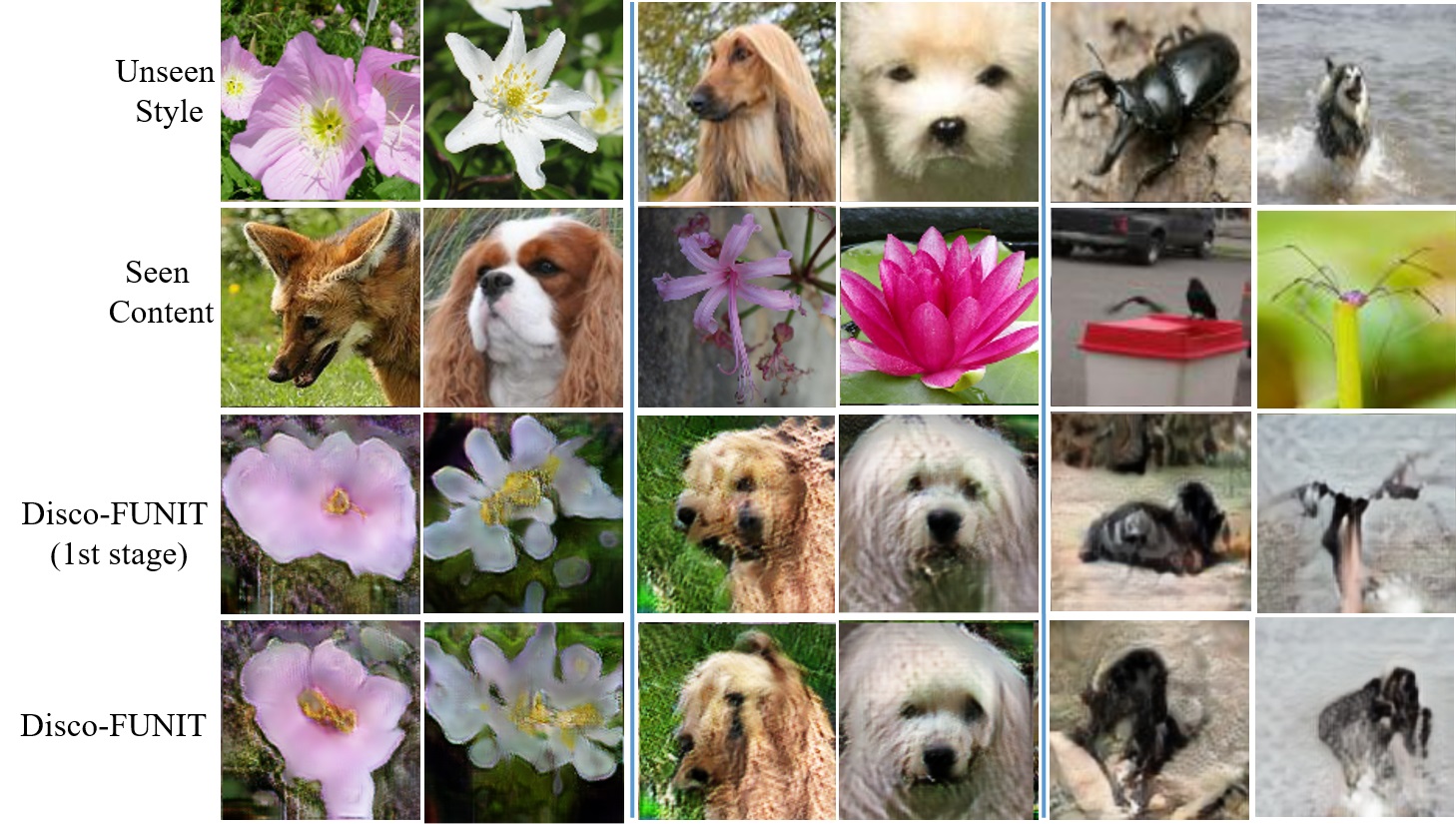}
\end{center}
\caption{Limitations of the proposed method. 1-2 (\emph{resp.,} 3-4, and 5-6) columns denotes the results of translating animal faces to flowers (\emph{resp.,} translating flowers to animal faces, and translating across categories from miniImagenet dataset). From top to bottom: unseen style images, seen content images, the images generated by Disco-FUNIT (1st stage), and the images generated by Disco-FUNIT by sampling a content map. 
}
\label{fig:limitation} 
\end{figure}

\section{Limitation of Our Method}\label{sec:limitation}
\noindent\textbf{Transfer Across Coarse-grained Categories}
Although our model achieves promising results both qualitatively and quantitatively on three real datasets, it fails to translate content image to style image when the category variance is significantly huge between content images and style images. (e.g. translating flowers (\emph{resp.,} animals) to animals (\emph{resp.,} flowers), translating across categories from challenging miniImagenet dataset~\cite{vinyals2016matching} with large inter-category variance). 
By taking flower to animals translation as an example, we randomly select content images from Flowers dataset and style images from Animal Faces dataset to learn a dictionary of local content vectors and disentanglement framework in the first stage. Then, we model the conditional distribution of discrete content maps based on real images from Flowers dataset.
In Figure~\ref{fig:limitation}, we show some examples generated by Disco-FUNIT (1st stage) and Disco-FUNIT. We can see that images generated by Disco-FUNIT (1st stage) roughly retain the content information of the content image as well as the appearance of the style image, but they are a little vague and lose details of style image. Since our content map generator is built on the learned disentanglement framework and the dictionary of local content vectors in the first stage, poor results in the first stage lead to unsatisfactory results in the second stage. In fact, existing few-shot image translation methods~\cite{liu2019few,saito2020coco,wang2020semi} and few-shot image generation methods~\cite{hong2020deltagan,hong2020f2gan} also fail to achieve transfer learning across coarse-grained categories. 

\noindent\textbf{Time Cost of Context Map Generator}
Our content map generator $T_{c|s}$ in the second stage adopts transformer architecture, which is time-consuming to sample 
local content vectors to form content maps autoregressively. 
We calculate the time cost of each module in our method at test time: 0.0009 seconds for the style encoder $E_s$, 6.5281 seconds for the content map generator, 0.0016 seconds for looking up the dictionary to construct content map, and 0.0031 seconds for the decoder $F$. We can see that the majority of time cost is from the content map generator. However, 
it is a common issue when using autoregressive models for image generation~\cite{yu2021diverse,esser2021taming,chen2018pixelsnail,razavi2019generating,peng2021generating}, instead of the unique problem of our method. 
We may consider employing more efficient Transformer or incremental sampling technique~\cite{ramachandran2019fast} to reduce the time cost of autoregressive models, which is out of the scope of this paper and will be studied in the future work. 




\bibliographystyle{ACM-Reference-Format}
\balance
\bibliography{egbib}